\newcommand{\CLASSINPUTbaselinestretch}{0.98}
\documentclass[10pt, journal]{IEEEtran}
\usepackage[utf8]{inputenc}
\usepackage[utf8]{inputenc}
\usepackage{parskip}
\usepackage{graphicx}
\usepackage[algo2e]{algorithm2e} 
\usepackage{algorithm2e}
\usepackage{caption}
\usepackage{algorithm}
\usepackage{algorithmic}
\usepackage{amsmath}
\usepackage{wrapfig}
\usepackage[export]{adjustbox}
\usepackage[noadjust]{cite}
\usepackage{cite}
\usepackage{xcolor}
\usepackage{soul}
\usepackage{microtype}
\usepackage[bottom]{footmisc}
\pdfoutput=1

\usepackage{graphicx}
\usepackage{subfigure}
\usepackage{booktabs} 

\allowdisplaybreaks
\title{Sparsity Turns Adversarial: Energy and Latency Attacks on Deep Neural Networks}

\begin{document}

\author{\IEEEauthorblockN{Sarada Krithivasan, Sanchari Sen and Anand Raghunathan} \\
\IEEEauthorblockA{School of Electrical and Computer Engineering, Purdue University \\ 
\{skrithiv, sen9, raghunathan\}@purdue.edu}
\thanks{This work was supported by C-BRIC, one of six centers in JUMP, a Semiconductor Research Corporation (SRC) program, sponsored by DARPA.} 
\thanks{Manuscript received April 18, 2020; revised June 12, 2020; accepted July 6, 2020. This article was presented in the International Conference on Compilers, Architecture, and Synthesis for Embedded Systems 2020 and appears as part of the ESWEEK-TCAD special issue.}
\vspace*{-0.1in}}

\maketitle

\vspace*{-10pt}

\begin{abstract}

Adversarial attacks have exposed serious vulnerabilities in Deep Neural Networks (DNNs), causing misclassifications through human-imperceptible perturbations to DNN inputs. We explore a new direction in the field of adversarial attacks by suggesting attacks that aim to degrade the energy or latency of DNNs rather than their classification accuracy. As a specific embodiment of this new threat vector, we propose and demonstrate adversarial sparsity attacks, which modify a DNN’s inputs so as to reduce sparsity (or the incidence of zeros) in its internal activation values. Exploiting sparsity in hardware and software has emerged as a popular approach to improve DNN efficiency in resource-constrained systems. The proposed attack therefore increases the execution time and energy consumption of sparsity-optimized DNN implementations, raising concern over their deployment in latency and energy-critical applications.

We propose a systematic methodology to generate adversarial inputs for sparsity attacks by formulating an objective function that quantifies the network’s activation sparsity, and minimizing this function using iterative gradient-descent techniques. To prevent easy detection of the attack, we further ensure that the perturbation magnitude is within a specified constraint and that the perturbation does not affect classification accuracy. We launch both white-box and black-box versions of adversarial sparsity attacks on image recognition DNNs and demonstrate that they decrease activation sparsity by 1.16x-1.82x. On a sparsity-optimized DNN accelerator, the attack results in degradations of 1.12x-1.59x in latency and 1.18x-1.99x in energy-delay product (EDP). Additionally, we analyze the impact of various hyper-parameters and constraints on the attack’s efficacy. Finally, we evaluate defense techniques such as activation thresholding and input quantization and demonstrate that the proposed attack is able to withstand them, highlighting the need for further efforts in this new direction within the field of adversarial machine learning.

\end{abstract}

\section{Introduction}
The widespread success of Deep Neural Networks (DNNs) in various machine learning applications, including image recognition, speech recognition, and natural language processing, has led to their deployment in several real-world products and services~\cite{DRL, DeepSpeech, Hinton}. State-of-the-art DNNs place immense computational and memory demands on the underlying computing platforms on which they execute. Consequently, a wide range of algorithmic, software and hardware techniques have been proposed to improve the execution efficiency of DNNs. 

\begin{wrapfigure}{r}{0.67\columnwidth}
\centering
 \includegraphics[width=0.67\columnwidth,scale=1]{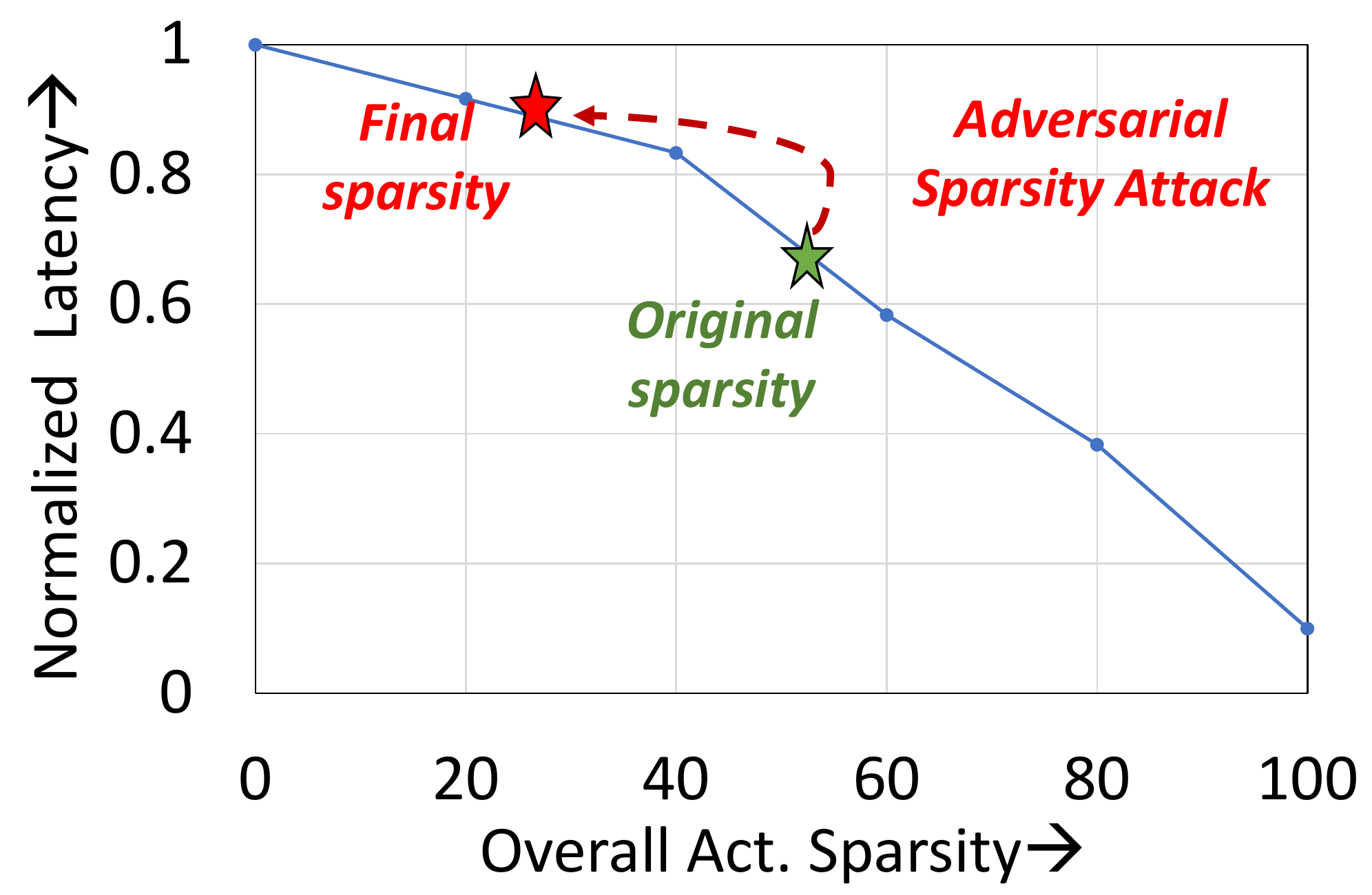}
 \caption{ Impact of activation sparsity on latency}
 \label{fig:VarIm}
\vspace*{-6pt}
\end{wrapfigure}

Several recent efforts optimize DNN implementations by exploiting \emph{sparsity}, or the prevalence of zero values in DNN weights and activations, to optimize storage and computation. For example, sparsity-optimized DNN implementations may store the sparse data-structures in a compact  format to reduce the memory requirements~\cite{DeepCompression} and/or skip redundant multiply-accumulate operations caused by zero-valued operands~\cite{SCNN,Cnvlutin,eyerissv2,Sparce,sparten,extensor, sparseTC}. We observe that on sparsity-optimized platforms, the execution time and energy consumption of a DNN strongly depend on the amount of sparsity present in the network. Figure~\ref{fig:VarIm} depicts the variation in latency of a sparsity-aware hardware accelerator \cite{ Cnvlutin} when evaluated on the ImageNet dataset for the VGG16 \cite{VGG19} network. As can be seen, increasing (decreasing) the activation sparsity leads to proportional decreases (increases) in latency.

We leverage the aforementioned property to expose a new vulnerability in sparsity-optimized DNN implementations. Specifically, we present an attack  that  \textit{adversarially perturbs the input so as to significantly reduce sparsity} in the internal activations of the network. This greatly diminishes or eliminates the benefits of sparsity optimizations, and leads to \textit{unanticipated increases in inference latency and energy consumption}. The impact of such an attack can range from disruptive to catastrophic, depending on the application.
For example, an autonomous self-driving car is expected to detect obstacles or unforeseen changes in the environment within around 600 milliseconds \cite{self_driving_car}. 
In that time, the Cnvlutin accelerator can process nearly 30 frames using the VGG16 DNN, assuming an average activation sparsity of 45-50\%. However, under the influence of the attack (detailed results are presented in Section~\ref{sec:expResults}), the average activation sparsity decreases by 1.79$\times$ and the latency of the accelerator is increased by nearly 1.4$\times$, causing it to process only 21 frames in the available time. The DNN could thus miss potentially critical data present in the remaining frames. For ensuring correct real-time operation, such an attack would require designers to over-design systems, essentially losing the benefits from sparsity optimization. The increased energy consumption resulting from sparsity attacks could also potentially lead to pre-mature battery discharge in battery-powered devices.

To realize adversarial sparsity attacks, we propose a systematic methodology that extends the principles of conventional adversarial attacks to identify the required input perturbations for any given DNN and input. We consider both white-box and black-box attack scenarios, which differ in the extent of the attacker's knowledge of the DNN model. In a white-box scenario, where the attacker has full access to the model parameters, we define an objective function that measures the activation sparsity for a given input. We then utilize gradient-descent-based techniques to identify the required distortions that shift the input in the direction of decreasing the sparsity objective function. In black-box scenarios wherein model information is unavailable to the attacker, we demonstrate transferability of sparsity-attacked inputs. Further, in both cases, the attack is designed to be functionally inconspicuous and cannot be detected by mere inspection of the DNN's input and output.
In summary, the attack methodology transforms an input into a similar input that has no net functional effect on the classification, but incurs significantly reduced sparsity within the DNN and hence, increased classification time and energy.

We analyze the effectiveness of the attack by studying its performance under common defense techniques including input quantization and activation thresholding. Despite attempting to counter the attack's impact by increasing activation sparsity, we find that these defenses are unable to recover sparsity without sacrificing the accuracy of the network. This underscores the strength of the attack and also the need for future research efforts in developing more effective defense techniques. Finally, we also establish that the attack is portable across different sparsity-optimized hardware platforms, including a specialized accelerator and a general-purpose processor.


In summary, the key contributions of this work are:
\vspace{-2pt}
\begin{itemize}
    \item We introduce adversarial sparsity attacks, a new class of attacks that affect the latency and energy consumption of DNNs on sparsity-optimized platforms. 
    \item We propose systematic methodologies to introduce adversarial perturbations to DNN inputs in a manner that causes a decrease in activation sparsity without impacting classification accuracy. Our attack is applicable under both white-box and black-box scenarios.
    
    \item We launch the attack on a set of 4 DNNs across 3 different datasets (MNIST, CIFAR-10 and ImageNet), and demonstrate a 1.16$\times$-1.82$\times$ decrease in activation sparsity with no loss in accuracy. We also demonstrate 1.12$\times$-1.59$\times$ and 1.18$\times$-1.99$\times$ increase in execution time and energy-delay product on a sparsity-optimized DNN accelerator. 
    \item We investigate the effectiveness of three different defense techniques and demonstrate that the attack is largely resistant to these defenses, highlighting the need for further efforts towards effective defenses.
    
\end{itemize}

The rest of the paper is organized as follows. Section~\ref{sec:prelim} provides the necessary background on sparsity in DNNs as well as conventional adversarial attacks. Section~\ref{sec:design} describes sparsity attacks and the proposed attack generation framework, and Section~\ref{sec:defense} discusses potential defense techniques. The experimental methodology used to evaluate sparsity attacks is described in Section~\ref{sec:exptMethod}. Section~\ref{sec:expResults} presents the results of applying sparsity attacks to DNN benchmarks. Finally, Section~\ref{sec:conclusion} concludes the paper.

\section{Background and Related Work}
\label{sec:prelim}
This section describes the different types of sparsity in DNNs and their sources. It also outlines the basic principles of conventional adversarial attacks that are utilized in the proposed sparsity attack framework.
\vspace{-8pt}
\subsection{Sparsity in DNNs}
Sparsity refers to the presence of zero-valued elements in different DNN data structures like weights and activations. Sparsity helps reduce the memory requirements of DNNs through the use of compact sparse storage formats. In addition, multiply-and-accumulate operations on zero-valued operands are redundant, and skipping them helps reduce time and energy consumption in DNN implementations~\cite{SCNN,Cnvlutin,eyerissv2,Sparce,sparten,extensor}.

Sparsity in DNNs can be broadly classified into two categories:
\begin{itemize}
    \item \textbf{Static Sparsity}: In static sparsity, the number and locations of zero values remain constant across different inputs to the network. Static sparsity arises in the network parameters, {\em viz.} weights, and is typically introduced through network pruning~\cite{DeepCompression}.

%
    
    \item \textbf{Dynamic Sparsity}: In dynamic sparsity, the number and locations of zero values vary across different inputs to the network. 
    Sparsity in activations is the primary example of dynamic sparsity and is caused by the presence of ReLU (Rectified Linear Unit) layers, which zero out all negative values. State-of-the-art DNNs are known to contain significant dynamic sparsity. For example, activation sparsity varies between 48-74\% among the DNNs used in our experiments.

\end{itemize}


\begin{wrapfigure}{r}{0.67\columnwidth}
\centering
 \includegraphics[width=0.67\columnwidth,scale=1]{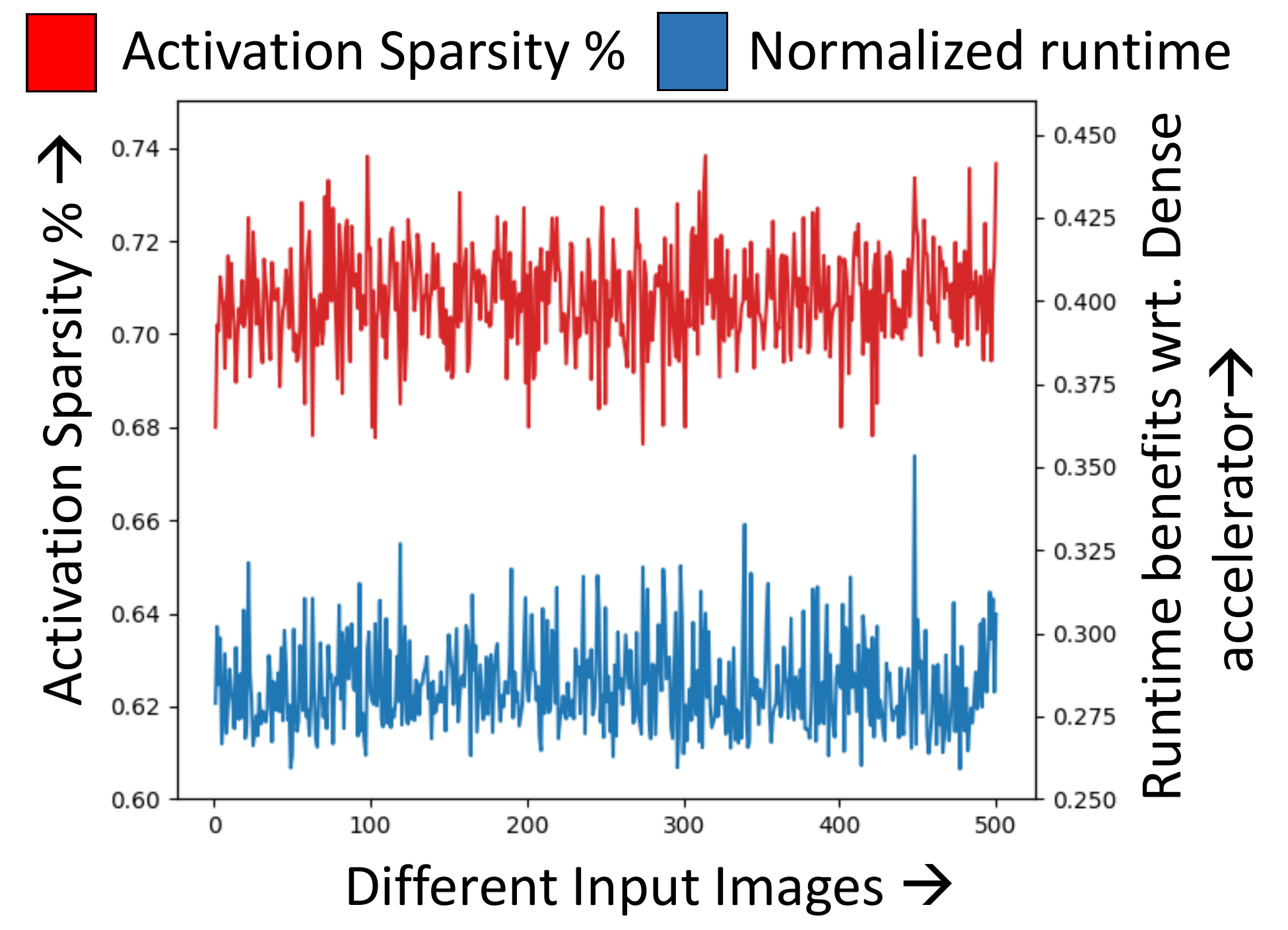}
 \caption{Variation in input activation sparsity and execution time benefits across 500 images for Cifar10-Conv2 on Cnvlutin}
 \label{fig:DynVar}
\vspace*{-2pt}
\end{wrapfigure}
Figure~\ref{fig:DynVar} illustrates the variation in dynamic sparsity across a set of 500 images from Cifar10-Conv2, a network trained on Cifar-10 \cite{cifar10}, with the architecture listed in Figure~\ref{table:network}. We observe a variation of nearly 6\% across the different images. The degree of sparsity in a DNN determines its execution time and energy consumption on sparsity-optimized platforms~\cite{SCNN,Cnvlutin,eyerissv2,Sparce,sparten,extensor, sparseTC}. Thus, the execution time of Cifar10-Conv2 varies by nearly 10\% for the same 500 images on the Cnvlutin accelerator platform~\cite{Cnvlutin}. Our proposal of adversarial sparsity attacks exploits this property to modify inputs to a DNN such that the amount of dynamic sparsity is reduced, thereby negatively affecting the DNN's execution time and energy consumption on sparsity-optimized platforms. 
\vspace{-0.12in}

\subsection{Adversarial Attacks on DNNs}
\vspace{-0.08in}
Adversarial attacks~\cite{IFGSM} have exposed serious security vulnerabilities in DNNs, raising a concern with their deployment in  safety-critical applications like autonomous cars, unmanned aerial vehicles and healthcare. 
To date, these attacks focus on causing a neural network to misclassify an input by introducing small perturbations (typically imperceptible to human eyes). 

Adversarial attacks can be classified as \textit{untargeted} or \textit{targeted}, depending on whether the aim of the attack is to generally degrade classification accuracy or to misclassify the input as a specific incorrect class. These attacks can be further classified on the basis of the extent of information available to the attacker about the DNN model. {\em White-box attacks} assume that the attacker has complete knowledge of the network structure and parameters in addition to the class labels. On the other hand, in {\em black-box attacks}, the attacker is only aware of the network outputs (class labels and/or confidence scores), but does not know the network structure and parameters. In many deployed systems, the fact that a DNN is used for classification, the network structure, or the entire model including weights are made public ({\emph e.g.}, DeepFace from Facebook \mbox{\cite{deepface}} and GPT3 from OpenAI \mbox{\cite{gpt3}}), allowing attackers to launch both white-box and black-box attacks.

White-box attacks such as C$\&$W\cite{CW} and MIM\cite{MIM} calculate the gradients of the classification loss function with respect to the input. This gradient information is used to compute the necessary input perturbations that increase the probability of mis-classification. 
Black-box attacks, however, can't utilize the same gradients in the absence of information about the model parameters. Popular black-box attack strategies usually overcome this difficulty by approximately estimating the gradients \cite{Zoo} or the required perturbation directions \cite{SimpleBB}. For example, ZOO \cite{Zoo} estimates the gradient of the classification probabilities at a particular point in the input space by querying the user model repeatedly to obtain the classification scores at several closely spaced input points. The gradient is constructed by calculating the change in the output classification scores observed for the very small changes in the input values around the desired point. 
Other common black-box strategies take advantage of the transferable nature of the adversarial inputs \cite{BB}.

Complementary to prior adversarial attacks, which aim to impact the DNN's accuracy, adversarial sparsity attacks instead affect the computational efficiency of DNNs, increasing their latency and energy consumption.


\subsection{Other Attacks on DNNs}
\vspace{-5pt}
We briefly elucidate some other forms of attacks on DNNs. The objective of timing side channel attacks, proposed in \cite{tsc}, is to more effectively launch black-box accuracy-based adversarial attacks or membership inference attacks \mbox{\cite{membership_inference}}, which discern whether a data sample belongs to the user's training dataset. The attack utilizes the execution time of a network to infer parameters like network depth, and constructs a new network that closely mimics the functionality of the unknown model. In fault injection attacks targeting DNNs \cite{bitflip, survey, fault_inj} the attacker causes severe misclassifications by merely modifying a small percentage of DNN parameters, such as by flipping certain carefully selected bits of the DNN model. However, the proposed attacks that have been demonstrated in this domain do not target or affect the computational efficiency of the model on a hardware platform and are thus orthogonal to the adversarial sparsity attacks proposed in our work.
 


The following sections describe the details of adversarial sparsity attacks.

\section{Adversarial Sparsity Attacks}
\label{sec:design}
Adversarial sparsity attacks are a new class of attacks that perturb DNN inputs so as to decrease the sparsity of activation values, with the eventual goal of increasing DNN execution time and energy on sparsity-optimized platforms. This section presents an overview of these attacks and details the proposed attack generation framework.
\vspace{-5pt}
\subsection{Threat Models}
\begin{figure}[t!]
\centering
\vspace*{-3pt}
 \includegraphics[width=\columnwidth,scale=1]{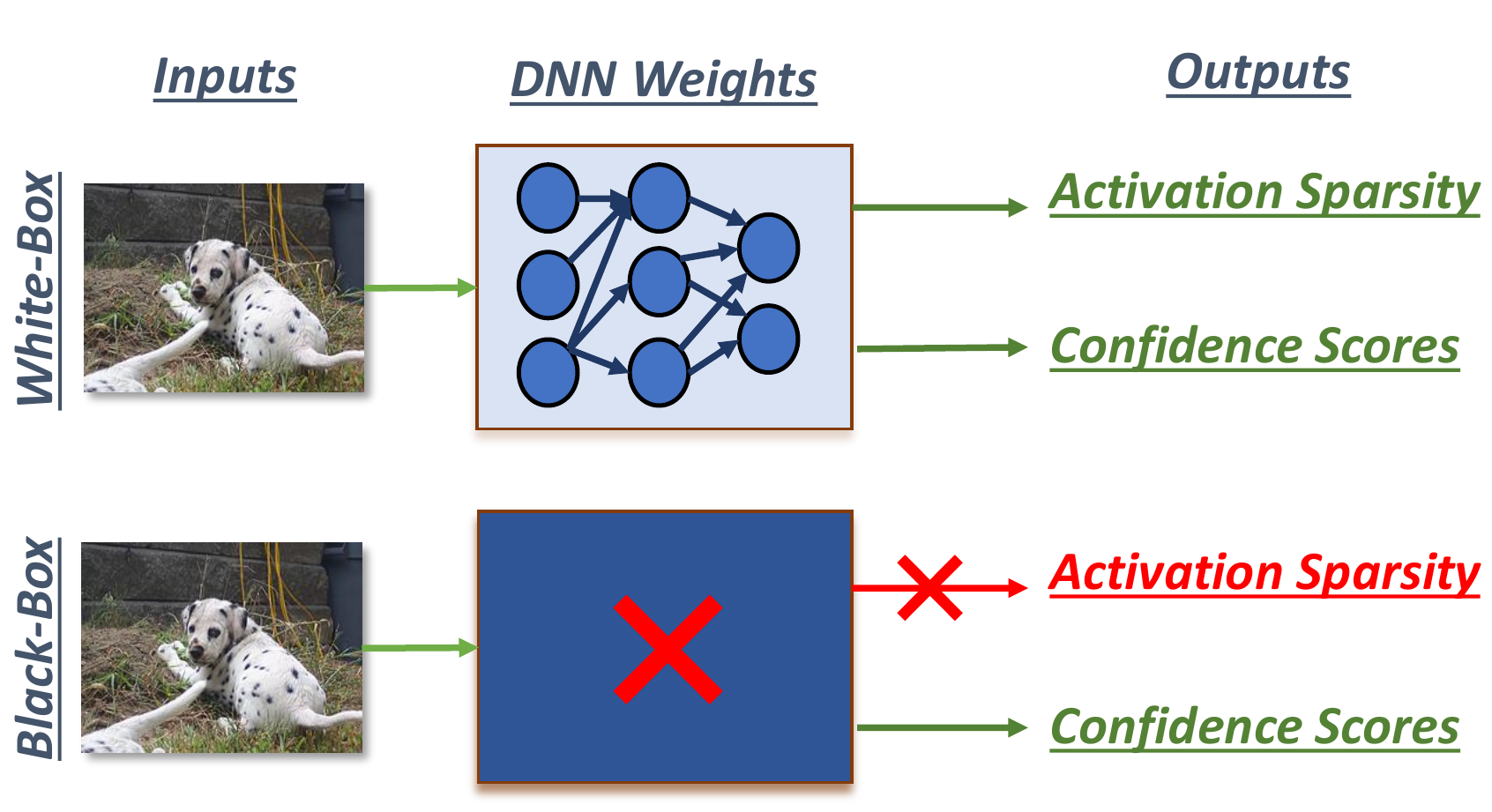}
 \vspace*{-11pt}
 \caption{Adversarial Sparsity Attacks in White and Black Box Settings}
 \label{fig:ThreatModel}
\vspace*{-6pt}
\end{figure} 
In adversarial sparsity attacks, the intent of the attacker is to induce high classification time and energy consumption through carefully crafted input perturbations. For a pronounced impact on the system's battery life and classification latency, the attack needs to be persistent, \emph{i.e.}, launched on several inputs over a long period of time. To facilitate this, the attack must evade trivial detection mechanisms such as observing the input and functional output of a DNN for anomalies. Specifically, the attack minimizes its impact on the input by adding perturbations that are bounded within a specified limit $L_{p}$, so that the resultant adversarial input is similar to the unperturbed input. In addition, the attack should ensure that the classification label assigned to an adversarial input is identical to that assigned to the original input and consequently, the functional output of the DNN remains unchanged. In other words, the attacker must craft an adversarial sample that is assigned the same class label, regardless of the correctness of the original classification. Overall, the criteria for a successful sparsity attack can be summarized as: 


\begin{enumerate}
    \item The added perturbation must cause a significant reduction in activation sparsity
    \item To prevent detection of the attack by simply observing the DNN output, the classification prediction must be the same as that of the unperturbed input
    \item To prevent detection of the attack by visual inspection of the input, the added distortions to the input must be imperceptible, \emph{i.e.}, within a specified bound $L_{p}$  
\end{enumerate}

Criterion 3 can be relaxed in scenarios wherein there is no supervision of the input image, such as autonomous navigation \cite{Drone}. In such scenarios, the only feedback available to the user is the system behavior ({\em e.g.}, credibility or accuracy of the navigation), which is ensured to remain the same by satisfying Criterion 2.

We now discuss different attack scenarios that place varying constraints on the attacker's knowledge of the user model. Figure~\ref{fig:ThreatModel} illustrates how sparsity attacks can be launched in both white-box and black-box scenarios. The attacker in a white-box scenario has full access to the user model's parameters such as the weights, and can consequently determine the internal activations. The attacker can thus calculate the overall activation sparsity for a given input, as well as its classification performance. In contrast, the attacker in a black-box scenario cannot access the model's internals, including weights and activations. Similar to state-of-the-art black-box attacks \cite{Zoo, SimpleBB} we assume that the attacker can present an input to the DNN and obtain the output class label and confidence scores. However, the attacker is prevented from evaluating the overall activation sparsity for a given input.

\vspace{-0.15in}
\subsection{Attack Strategies}
We first describe the attack strategies developed to realize sparsity attacks in white-box scenarios, and then address how they can also be applied to black-box scenarios.
\vspace{7pt}
\subsubsection{White-Box Sparsity Attacks}
\label{subsec:whiteboxstrategy}
The proposed white-box attack formulates a composite objective function that combines the overall activation sparsity with the classification loss of the network for a given input. Next, it calculates the gradient of this objective function with respect to the input to reveal the relationship between changes in the input and the corresponding effects on activation sparsity and classification loss. Finally, by incorporating this gradient into iterative gradient-descent optimizers, it obtains the necessary input perturbations that decrease activation sparsity while minimizing the classification loss. 


The objective function, $\mathcal{L}$, that combines the overall activation sparsity and classification loss for a given input is described by Equation~\ref{eq:objective}.
\begin{equation}
\label{eq:objective}
    \mathcal{L}(x) = \mathcal{L}_{sparsity}(x) + c * \mathcal{L}_{ce}(x)
\end{equation}

where $c$ is the Lagrange multiplier or the trade-off constant, $\mathcal{L}_{sparsity}$ estimates the overall activation sparsity across the network for an input $x$, $\mathcal{L}_{ce}$ measures the classification cross-entropy loss of the input $x$ with respect to the target class assigned by the network on the original unperturbed input. Note that, this is unlike network training, where $\mathcal{L}_{ce}$ is calculated against the ground-truth labels. In contrast, the goal in the attack is simply to mimic the DNN's behavior on unperturbed inputs.

The intent of the attacker is to calculate input perturbations that minimize the objective function $\mathcal{L}$. The gradient of $\mathcal{L}$ with respect to the DNN input is thus used to calculate the required perturbations that push the input in the direction of decreasing activation sparsity, while maintaining the predicted labels to be identical to that of the original input. Intuitively, the choice of the trade-off constant $c$ impacts the strength of the sparsity attack. It is important to choose a value of $c$ that achieves a high degradation in activation sparsity without altering the DNN's functional behavior. Further, it must also be ensured that the added perturbations are within the maximum $L_{2}$ distortion allowed, \emph{i.e.}, Criterion 3. The method used to derive adversarial inputs considering these factors is outlined in Algorithm~\ref{alg:ASA_im}. 

\begin{algorithm}[t]
\caption{Creating adversarial inputs for white-box sparsity attacks}
\label{alg:ASA_im}
\begin{algorithmic}[1]
\renewcommand{\algorithmicrequire}{\textbf{Input:}}
\renewcommand{\algorithmicensure}{\textbf{Output:}}
\REQUIRE{$x_{clean}$ (Clean Input), $f$ (DNN model), $\mathcal{L}_{sparsity}$ and $\mathcal{L}_{ce}$ (Objective function terms),  $\epsilon$ (Maximum $L_{2}$ distortion), $\epsilon_{iter}$ ($L_{2}$ distortion per iteration), $O_{max}$ and $I_{max}$ (Maximum outer and inner-loop iterations), $c_{in}$, $c_{min}$ and $c_{max}$ (Initial, min. and max. value of trade-off constant)} 
\ENSURE {$x_{adv}$ (Adversarial input)}
\STATE {$c$ = $c_{in}$, $o$ = 1}
\STATE {{\bf while} $o$ $<$ $O_{max}$ }
\STATE {\text{ }\text{ }$x_{0}$ = $x_{clean}$, $i$ = 1, $g_{0}$ = 0}
\STATE {\text{ }\text{ }{\bf while} $i$ $<$ $I_{max}$}
\STATE {\text{ }\text{ }\text{ }\text{ }$\mathcal{L}(x_{i})$ = $\mathcal{L}_{sparsity}(x_{i})$ + $c * \mathcal{L}_{ce}(x_{i})$} 

\STATE {\text{ }\text{ }\text{ }\text{ }$g_{i}$ = $\mu$$g_{i-1}$ + $\nabla_{x}\mathcal{L}(x_{i})$} \label{eq:grad1}
\STATE {\text{ }\text{ }\text{ }\text{ }$x_{i+1}$ = $x_{i}$ - $\epsilon_{iter} * \frac{g_{i}}{||g_{i}||_{2}}$}
\STATE {\text{ }\text{ }\text{ }\text{ }$x_{i+1}$ = $Clip_{[0,1,\epsilon, x_{clean}]}$ ($ x_{i+1}$)}
\STATE {\text{ }\text{ }\text{ }\text{ }$i$ = $i+1$}
\STATE {\text{ }\text{ }{\bf if} $argmax(f(x_{adv}))$ $\neq$ $argmax(f(x_{clean}))$ }
\STATE {\text{ }\text{ }\text{ }\text{ }$c$ = ($c$ + $c_{max}$)/2}
\STATE {\text{ }\text{ }{\bf else}}
\STATE {\text{ }\text{ }\text{ }\text{ }$c$ = ($c$+$c_{min}$)/2}
\STATE {\text{ }\text{ }\text{ }$o$ = $o$+1}

\end{algorithmic}
\end{algorithm}

Algorithm~\ref{alg:ASA_im} is organized as two nested loops. In the inner loop (Lines 4-9), for a particular value of the trade-off constant $c$, the algorithm uses gradient-descent techniques to calculate the ($L_{2}$-bounded) perturbations that decrease activation sparsity while maintaining classification output. In the outer loop, the algorithm employs simple binary-search techniques that identify an optimal $c$. 

Focusing on the gradient-descent process first, we begin by initializing the perturbed input $x_{0}$ to the original input $x_{clean}$ (line 3). In every iteration $i$, the gradient of $\mathcal{L}(x_{i})$ with respect to the input is evaluated (line 5). There are several gradient-descent based optimization algorithms that can be incorporated to update the input. In our experiments, we find that utilizing stochastic gradient descent with a momentum-based update \cite{momentum}, as shown in line~\ref{eq:grad1}, provides good results. The momentum parameter $\mu$ is set to 0.9 in all our experiments. Using the gradient values, a small distortion is calculated so as to push $x_{i}$ in the direction of decreasing the objective function (line 7). Finally, at the end of iteration $i$, the perturbed input is clipped (if needed) so as to ensure that the added distortions are within the $L_{2}$ bound $\epsilon$, and does not exceed the permissible input value range i.e., [0, 1] (line 8). This process is continued for $I_{max}$ iterations. 

At the end of $I_{max}$ iterations, the inner loop terminates and the binary search process for finding an optimal $c$ comes into play. The classification prediction of the perturbed input is first evaluated and compared against the prediction on the original input (line 10) - if the labels do not match, the priority of the $\mathcal{L}_{ce}$ term is increased by increasing $c$ as indicated in line 11. On the contrary, if the labels do match, we assign the $\mathcal{L}_{sparsity}$ term a higher priority by decreasing $c$ (line 13). Starting the next outer-loop iteration the perturbation and gradient variables are reset(line 3), and the process is continued until $O_{max}$ iterations are complete. The output of the algorithm is an adversarially perturbed input $x_{adv}$ that successfully decreases the activation sparsity, while satisfying input perturbation and output constraints. 

Naturally, the convergence and the resulting performance of Algorithm~\ref{alg:ASA_im} is contingent on the specific nature of the $\mathcal{L}_{ce}$ and $\mathcal{L}_{sparsity}$ functions. We next describe the process of suitably designing these objective function terms that play a crucial role in determining the efficacy of the adversarial sparsity attack. 

\textbf{Design of $\mathcal{L}_{sparsity}$}: For successfully satisfying Criterion 1, the function representing $\mathcal{L}_{sparsity}$ must effectively quantify the overall activation sparsity across the network for a given input. Ideally, we could apply a step function on the input activations of every layer in the network to determine whether it is non-zero, and add the step-function outputs across the network to determine the total activation sparsity. However, the objective function must also be differentiable in the input activation range so that the gradients can be calculated in Line~\ref{eq:grad1} of Algorithm~\ref{alg:ASA_im}. To that end, we devise continuous approximations of the step function to detect non-zero activation values in the network. 

We specifically use Tanh and Sigmoid functions, described by the following equations, in our framework:
\begin{equation}
    Tanh(\beta, act) = \frac{e^{\beta \cdot act} - e^{-\beta \cdot act}}{e^{\beta \cdot act} + e^{-\beta \cdot act}} 
\end{equation}
\begin{equation}
    Sigmoid(\beta,act) = \frac{1}{e^{-\beta \cdot act}+1}
\end{equation}

where $act$ is an activation value from the network. Both the functions are differentiable, and the value of $\beta$ can be increased to improve their resemblance to the step function. An estimate of the number of non-zero activations across all the layers of the network is obtained by summing the activations with the tanh or sigmoid functions applied on top of them. We define such an estimate $E$ as:
\begin{equation}
    E(x) =  \sum_{L} \sum_{N_{l}} F(I(n_{l,x})) 
\end{equation}  
where $F$ refers to the sigmoid or tanh function and $I(n_{l,x})$ refers to the $n^{th}$ input activation of layer $l$ for a given input, $x$. To reflect the number of zero-valued activations, $E$ must be negated. We further normalize $E$ to the total number of neurons in the network. The final $\mathcal{L}_{sparsity}$ can thus be expressed as: 
\begin{equation}
     \mathcal{L}_{sparsity}(x) =  -\frac{E(x)}{k}
    \label{eq:Sparsity}
\end{equation}
where $k$ refers to the network's total neuron count.

\begin{wrapfigure}{r}{0.5\columnwidth}
\centering
\vspace*{-3pt}
 \includegraphics[width=0.5\columnwidth,scale=1]{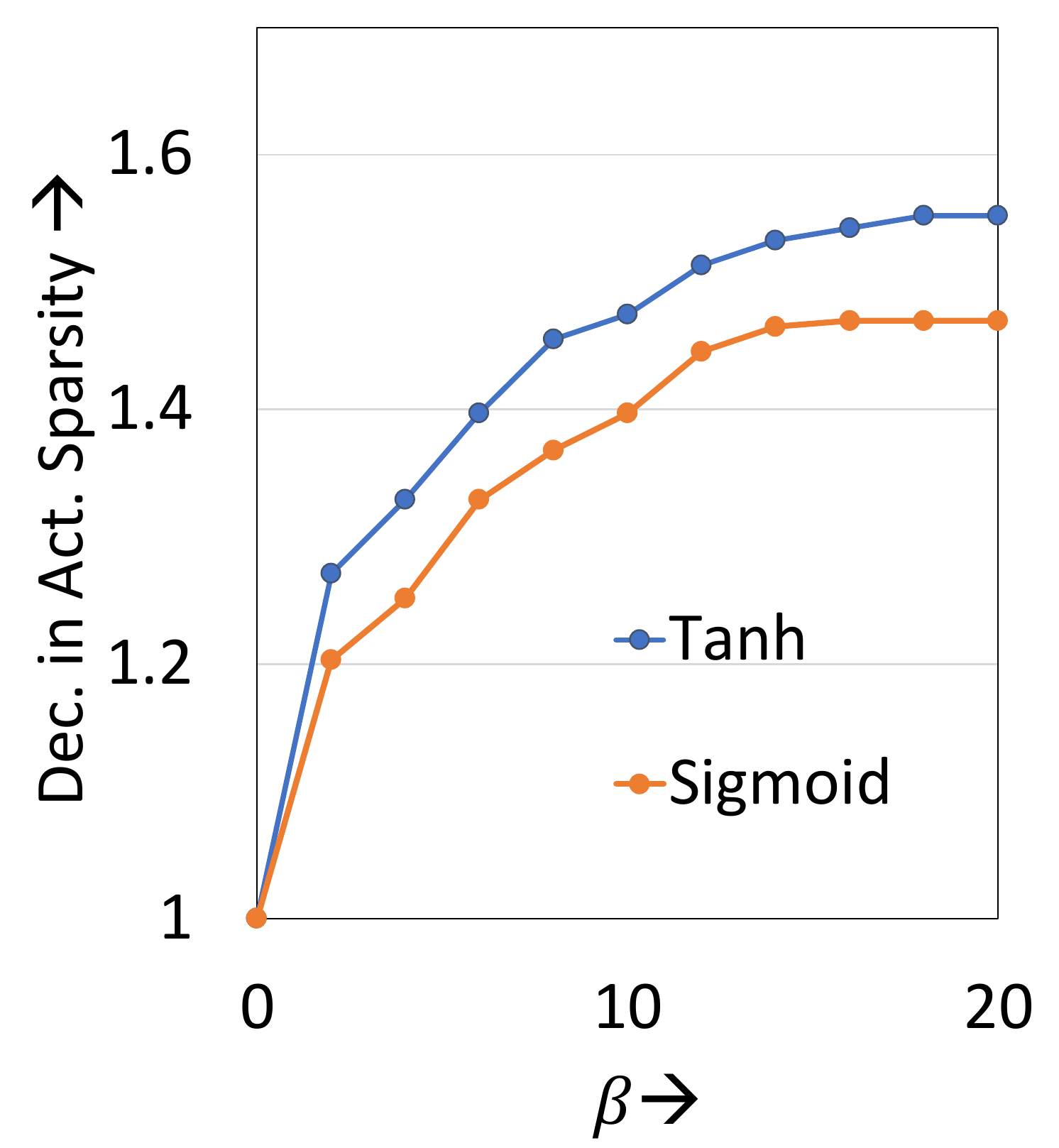}
 \caption{ Impact of $\beta$ on decrease in activation sparsity}
 \label{fig:VarBeta}
\vspace*{-6pt}
\end{wrapfigure}

Figure~\ref{fig:VarBeta} illustrates the efficacy of the sparsity attack on Cifar10-Conv2, using tanh and sigmoid objective functions with varying 
$\beta$. As can be seen for both the tanh and sigmoid functions, the decrease in activation sparsity obtained by the attack improves with increasing $\beta$ until saturation is reached at some point. Across all networks, for a particular value of $\beta$, we observe that the tanh function tends to perform better than the sigmoid function by decreasing the final activation sparsity by an additional 8\% on an average. We report the $\beta$ values used in conjunction with the tanh function for the best results on the networks considered in our experiments in Section~\ref{sec:exptMethod}. 

To ensure a higher impact of reduced sparsity on the execution time and energy consumption of different sparsity-optimized platforms, we also augment the objective function $\mathcal{L}_{sparsity}$ with some hardware-specific parameters. We specifically weigh the sparsity terms corresponding to each layer with values proportional to their relative execution time and energy consumption on a given hardware platform. In other words, time and energy intensive layers are assigned higher weight values in the overall objective function. The modified sparsity estimator function $E$ can thus be described by the following equation.

\begin{equation}
    E(x) = \sum_{L} W_{l}\sum_{N_{l}} F(I(n_{l,x})) \label{eq:WtSparsity}
\end{equation} 

where $W_{l}$ is the weight assigned to layer $l$ and is set to be equal to the relative runtime or energy consumption of that layer on the  considered hardware platform. 

\textbf{Design of $\mathcal{L}_{ce}$}: As per Criterion 2, the classification label predicted by the network on the perturbed input must match the label assigned by the network on the unperturbed input. Accordingly, $\mathcal{L}_{ce}$ must measure the classification loss of the input with respect to the original predicted label. Akin to the design of objective functions for targeted attacks \cite{MIM}, we define $\mathcal{L}_{ce}$ as the softmax cross-entropy loss with respect to the `target' label $y$. For original and adversarial inputs $x_{clean}$ and $x_{adv}$ this can be expressed as:
\begin{equation}
    y = argmax(f(x_{clean})) 
\end{equation}
\begin{equation}   
    \mathcal{L}_{ce}(x_{adv}) = -log(\frac{e^{f(x_{adv})_{y}}}{\sum_{l=1}^{N} e^{f(x_{adv})_{l}}})
\end{equation}

where $f(x)$ is the pre-softmax output of the DNN for input x, $y$ is the class label assigned by the network on the original input $x_{clean}$, $N$ is the number of classes and $f(x)_{l}$ is the pre-softmax output with respect to class $l$. 


In summary, white-box sparsity attacks exploit their knowledge of the model parameters by utilizing iterative gradient-descent based techniques to find the required perturbations that satisfy Criteria 1-3.

\vspace{7pt}

\subsubsection{Black-Box Sparsity Attacks}
In this subsection, we describe strategies to launch a sparsity attack in a black-box scenario. 

As shown in Figure~\ref{fig:ThreatModel}, the attacker in a black-box scenario can access only the network output and is unaware of the network weights and internal activations. This precludes the attacker from directly utilizing the gradient-descent-based attack methodology described in Algorithm~\ref{alg:ASA_im}. As mentioned in Section~\ref{sec:prelim}, black-box attacks work around this challenge by iteratively determining approximations of either the gradient or the classification output \cite{Zoo}, or the appropriate direction of perturbation \cite{SimpleBB}. These approximations solely require the classification confidence scores for a given input, and do not require any knowledge of the network weights. Unfortunately, we cannot adopt similar techniques in black-box sparsity attacks as the amount of sparsity in a network cannot be estimated from the queried classification scores of the user model. Hence, we explore alternative techniques for launching a black-box sparsity attack.

\begin{figure}[t]
\centering
 \includegraphics[width=\columnwidth,scale=1]{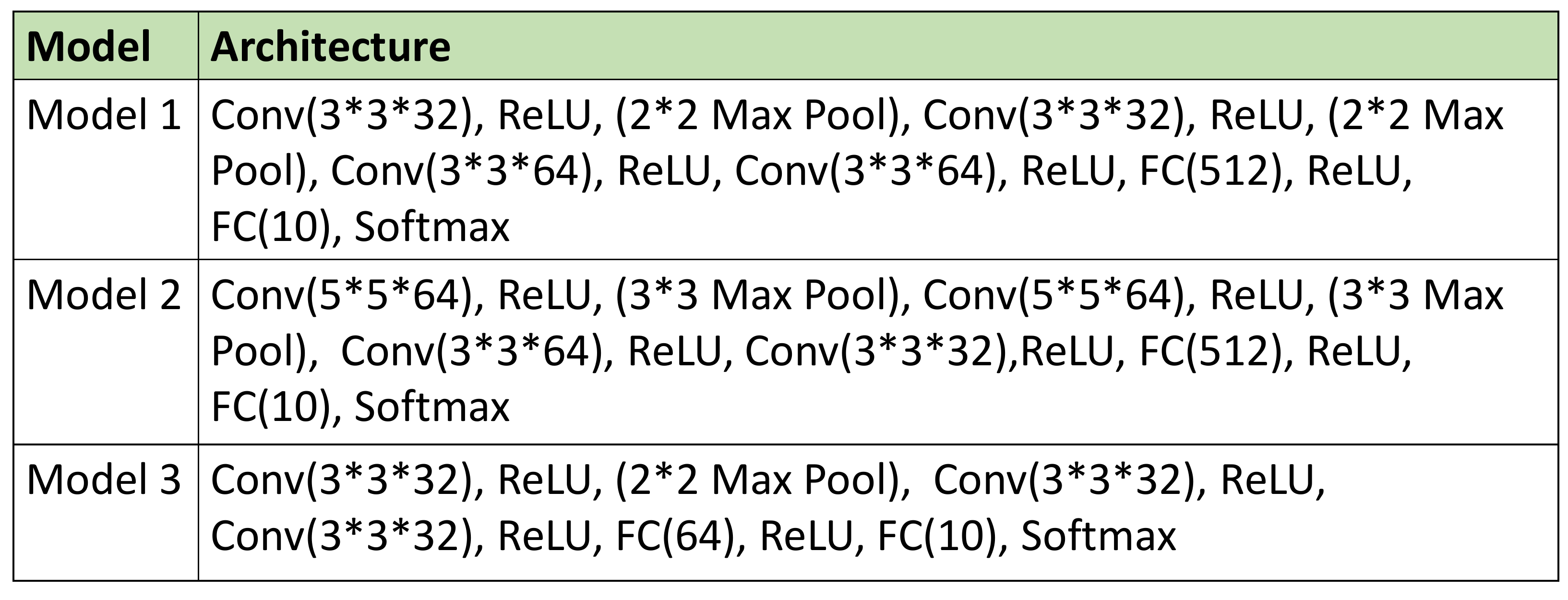}
 \caption{Architecture of different unknown models}
 \label{fig:VarBB_tab}
\vspace*{-8pt}
\end{figure}

First, we investigate whether adversarial sparsity attack inputs are \textit{transferable} between different networks, \emph{i.e.}, whether an adversarial input generated for a known substitute (or surrogate) model~\cite{BB} using the white-box technique can have the desired effects on the unknown user model. To evaluate such transferability, we first perform the white-box attack described in the previous subsection on Cifar10-Conv2 (the substitute model, and whose architecture is listed in Section~\ref{sec:exptMethod}), and subsequently transfer the images to 3 different Cifar10 models whose architectures are listed for reference in Figure~\ref{fig:VarBB_tab}, but assumed to be unknown to the attacker. 

\begin{wrapfigure}{r}{0.57\columnwidth}
\centering
 \includegraphics[width=0.57\columnwidth,scale=0.5]{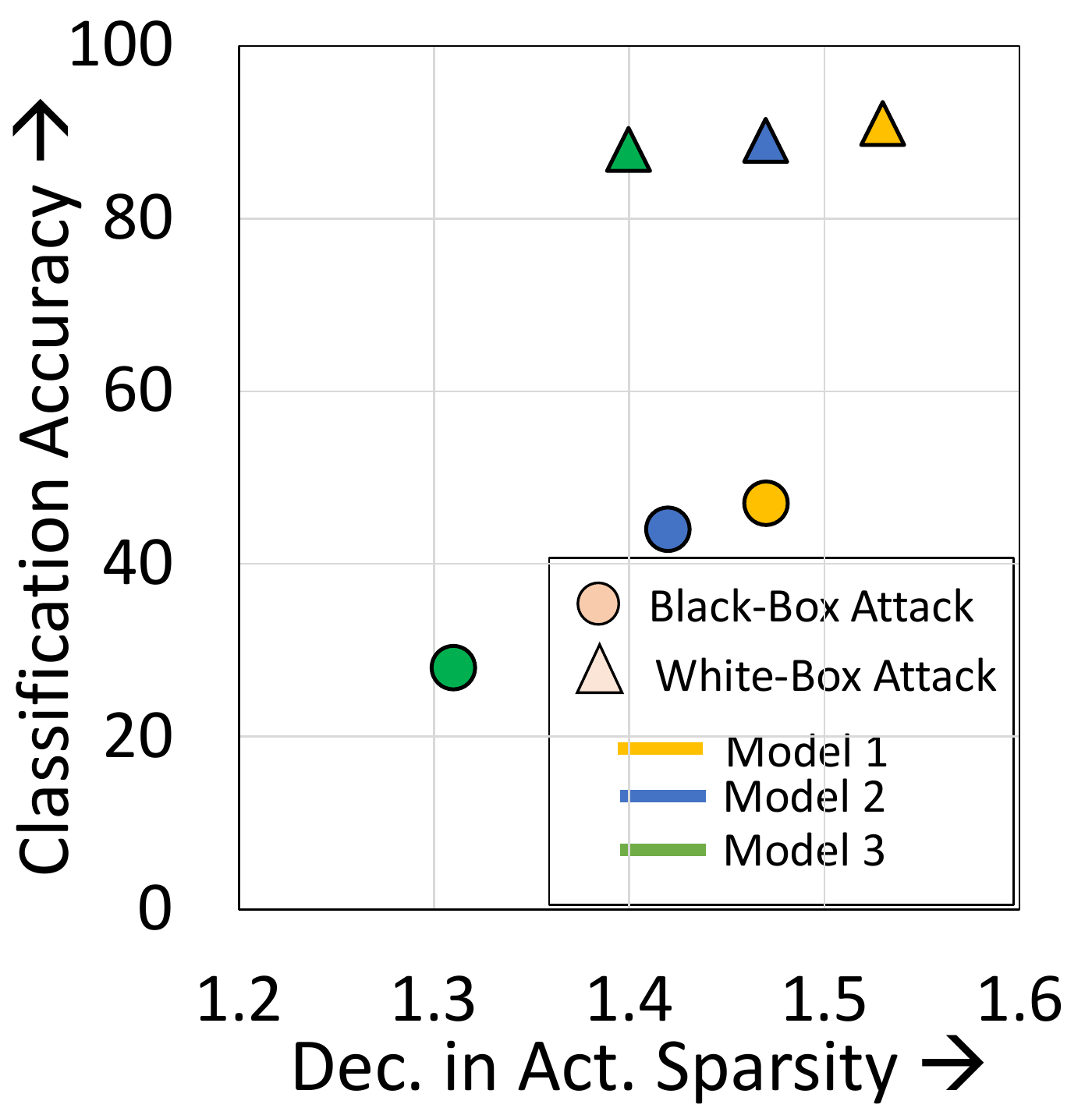}
 \caption{ Analyzing the transferability of the adversarial sparsity attack}
 \label{fig:VarBB}
\end{wrapfigure}

On deploying the white-box attack on Cifar10-Conv2, the average decrease in activation sparsity across the test dataset is 1.56$\times$, for no loss in classification accuracy. In Figure~\ref{fig:VarBB} we depict the decrease in activation sparsity and classification accuracy incurred by the unknown models when these images are passed through them. For the sake of comparison, we also illustrate the performance of the white-box sparsity attack on all three models. Across all unknown models, it is observed that the decrease in activation sparsity exhibited by simply transferring the images is only 10-12\% lower than that obtained by the white-box attack 
on the same model. However, unfortunately, there is a significant decrease in classification accuracy - nearly 60\%. Hence, while sparsity-attacked images appear to be transferable in terms of the reduction in activation sparsity, classification accuracy is clearly not maintained. We find this to be inline with the results reported in several accuracy-based black-box attack research efforts such as \cite{Zoo, BB_target}, which observe that unlike untargeted attacks, targeted attacks exhibit poor transferability, and require specialized techniques such as \cite{Zoo, SimpleBB, BB_target} for achieving high attack success rates.

\begin{figure}[t]
\centering
\vspace*{-11pt}
 \includegraphics[width=\columnwidth,scale=1.]{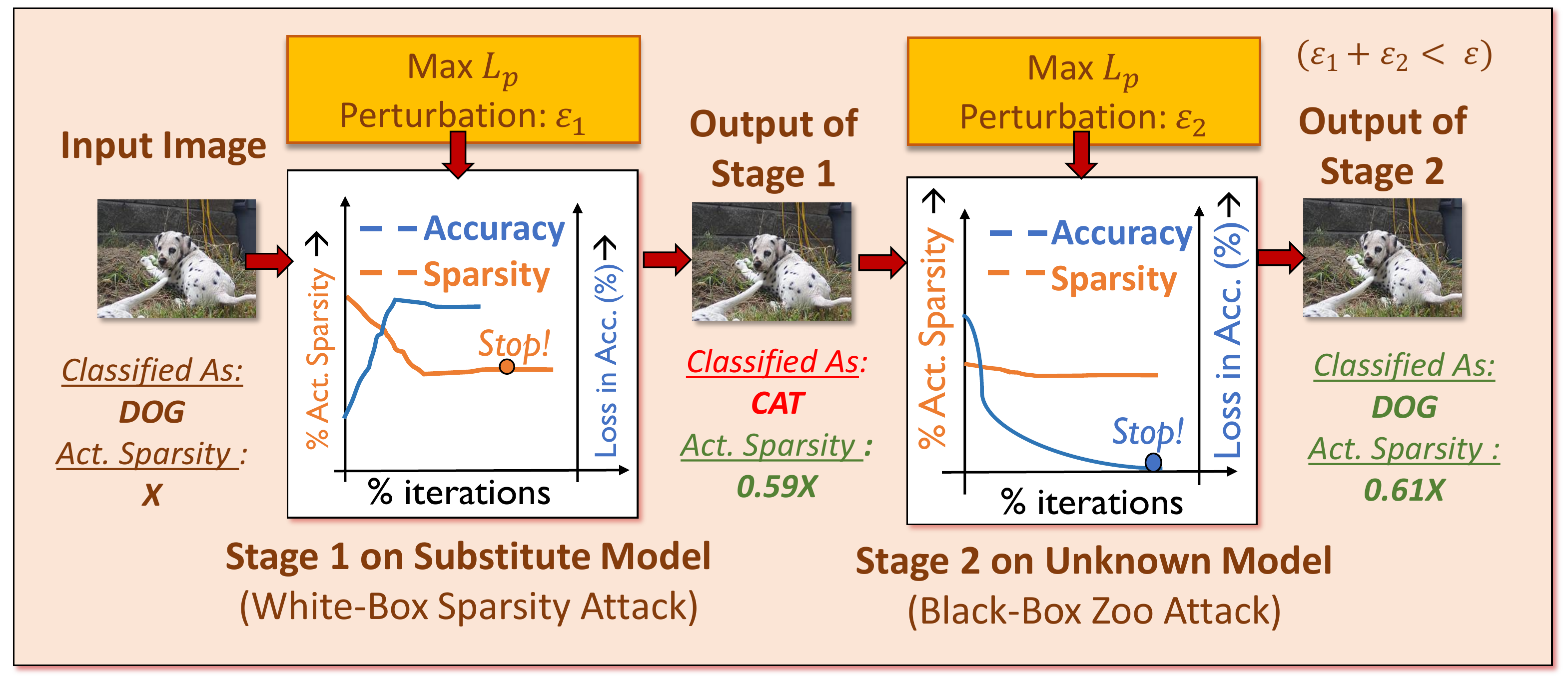}
 \vspace*{-11pt}
 \caption{Two Stages of the Black-Box Sparsity Attack}
 \label{fig:BB_2Stage}
\vspace*{-7pt}
\end{figure}

To resolve the aforementioned challenges, we propose a two-stage framework that first derives attack inputs on a known substitute model, transfers those inputs to the unknown model only for the purpose of reducing sparsity and (if needed) finally applies an accuracy-based targeted black-box attack to restore the original
prediction label. Figure~\ref{fig:BB_2Stage} summarizes the overall black-box sparsity attack process. As shown in  Figure~\ref{fig:BB_2Stage}, the first stage generates adversarial sparsity images on a substitute model using the white-box attack mechanism of Algorithm~\ref{alg:ASA_im}, with $c$ set to zero. In the second stage, to ensure that predictions are maintained, all mis-predicted images are passed through the ZOO targeted accuracy-based black-box attack~\cite{Zoo}. ZOO repeatedly queries and observes the changes in classification performance for some closely-spaced input points to determine the gradients for a particular input and utilizes these gradients to find the necessary perturbations to change the network output to the targeted class. As in white-box sparsity attacks, the target class is the class predicted by the unknown network on the unperturbed input. Experimental results indicate that although the second stage does not consider activation sparsity, its effects on sparsity are minimal (\textit{within 2\%} across the networks considered). In other words, the second stage does not undo the sparsity reduction obtained by the first stage. We also place additional constraints to ensure that the total $L_{2}$ distortion introduced by both stages ($\epsilon_{1} + \epsilon_{2}$) is within the specified limit $\epsilon$.

In summary, our black-box sparsity attack is realized in two stages: first transfer adversarial sparsity attack inputs created on a substitute model, and then apply targeted black-box attacks to restore the functional output if needed.

\section{Potential Defense techniques}
\label{sec:defense}

In this section, we outline potential defense techniques against adversarial sparsity attacks and evaluate their efficacy. The defense techniques broadly aim to restore activation sparsity, thereby countering the impact of the sparsity attack. In the process, successful defenses should not negatively impact the classification accuracy of the DNN as well. It is further imperative that the chosen defense be computationally lightweight --- it would be counter-productive if the defense itself added a significant latency or energy overhead in order to prevent the latency or energy increase incurred by the attack.

We specifically discuss the key principles of three different computationally light-weight defense techniques below. 


\vspace{-7pt}
\subsection{Activation Thresholding}

Activation thresholding, as the name suggests, involves setting all activations below a certain non-zero threshold to zero, thereby increasing the overall activation sparsity. The thresholds must be set to values that cause no loss in accuracy on unperturbed inputs and can be identified at a per-layer granularity. In our implementation, we initialize the threshold of each layer to the mean activation value of that layer as measured on the adversarial set, and iteratively increase or decrease its value till an increase in activation sparsity is obtained for a tolerable loss in accuracy.

\vspace{-4pt}
\subsection{Adversarial Training}
Adversarial training \cite{advTraining} is a popular defense technique used in classical accuracy-based adversarial attacks. It involves generating adversarial examples on the clean training set, and then retraining the model with the original training examples augmented with the adversarially generated samples. Such techniques have shown great success at mitigating the impact of attacks like FGSM \cite{FGSM}, PGD \cite{Madry}, \emph{etc}. 

Although adversarial training was formulated as a defense technique to mitigate accuracy-based attacks, we investigate its suitability for adversarial sparsity attacks. The aim of adversarial training in this context is to increase the activation sparsity of the adversarial inputs to match that of unperturbed input samples. We generate the inputs for adversarial training using the white-box techniques discussed in Section~\ref{sec:design}, and the parameters listed in Section~\ref{sec:exptMethod}.  
\vspace{-8pt}
\subsection{Input Smoothing}
Adversarial sparsity attack distortions are calculated perturbations added to the input so as to reduce activation sparsity while maintaining classification accuracy. We investigate whether the impact of the sparsity attack can be nullified by input smoothing techniques, namely input quantization and compression, similar to the techniques used in \cite{adv_defense} to defend against conventional adversarial attacks. We select appropriate parameters for each smoothing technique such that they mitigate the impact of the sparsity attack while incurring no loss in accuracy. For example, we select the bit-width or compression quality which provide the best defense against the attack. Input quantization to k bits applied on an input x is expressed as:
\begin{equation}
    x_{q} = \frac{1}{2^{k}} \cdot round((2^{k}) \cdot x)
\end{equation}

Input compression applies to images, and as the name suggests, consists of applying image compression to remove high-frequency artifacts. To realize input compression with quality $k$, we pass the input image $x$ to the $encode\_jpeg$ method from TensorFlow \cite{TF} with quality set to $k$.


For all of the above defenses, we assume that the attacker does not adapt to the defense technique. However, as we will be demonstrating in Section~\ref{sec:expResults}, none of the defenses explored here show significant mitigation of the sparsity attack - to reduce the impact of the sparsity attack by even 8\% (in activation sparsity) on the adversarial set, a 4\% loss in unperturbed accuracy must be incurred, which is an unacceptable tradeoff in most practical scenarios. This emphasizes the strength of the sparsity attack to withstand conventional defenses. We do not consider other popular techniques such as GANs \mbox{\cite{defenseGAN} or variational auto-encoders \cite{var_autoenc}} used in accuracy-based attacks as they incur significant computational overheads themselves. Likewise, ensembles of models \cite{empir, ensem_fp} have also been demonstrated to show improved robustness against adversarial attacks. Nevertheless, these may be interesting directions to investigate as part of future efforts.

A possible \textit{detection} mechanism involves measuring the latency or the energy consumption of the hardware system. If the energy consumption goes beyond a certain range, the user could be alerted. We note however that while this aids in detection of the attack, it does not resolve or counter the impact of the attack. Further research is thus needed to develop a successful defense technique.

\vspace*{1pt}
\section{Experimental Methodology}
\label{sec:exptMethod}
\begin{figure}[h!]
 \includegraphics[width=\columnwidth]{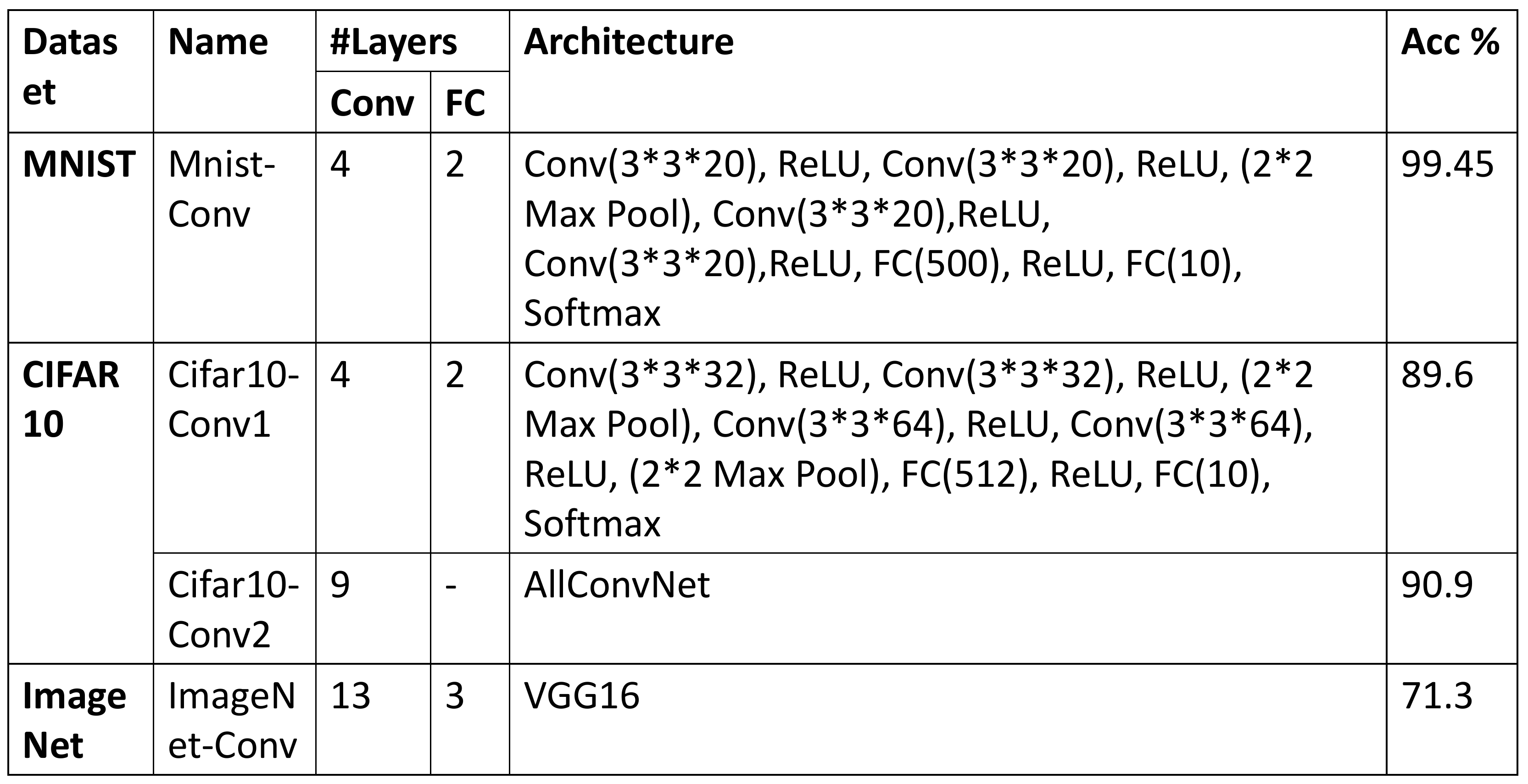}
 \caption{Employed model architectures}
 \label{table:network}
\end{figure}

\textbf{Datasets and Model Architectures}. We demonstrate the effectiveness of adversarial sparsity attacks on 4 different image-recognition DNNs across 3 different datasets (MNIST \cite{mnist}, CIFAR-10 \cite{cifar10} and ImageNet \cite{imagenet}). The details of the DNNs are listed in Figure~\ref{table:network}. The ImageNet-Conv and Cifar10-Conv2 architectures are taken from \cite{AllConvNet,VGG19}. The white-box and black-box attacks are conducted on the full test set, except in the case of the black-box attack launched on ImageNet-Conv wherein we report results for 1000 randomly selected test set images. The reduction in activation sparsity inflicted by the attack does not vary significantly across images ($<$5\% variation). We therefore report the average reduction in activation sparsity, and the corresponding increase in classification latency and energy-delay product, for each network in Section~\ref{sec:expResults}. 

\begin{figure}[h!]
  \includegraphics[width=\columnwidth]{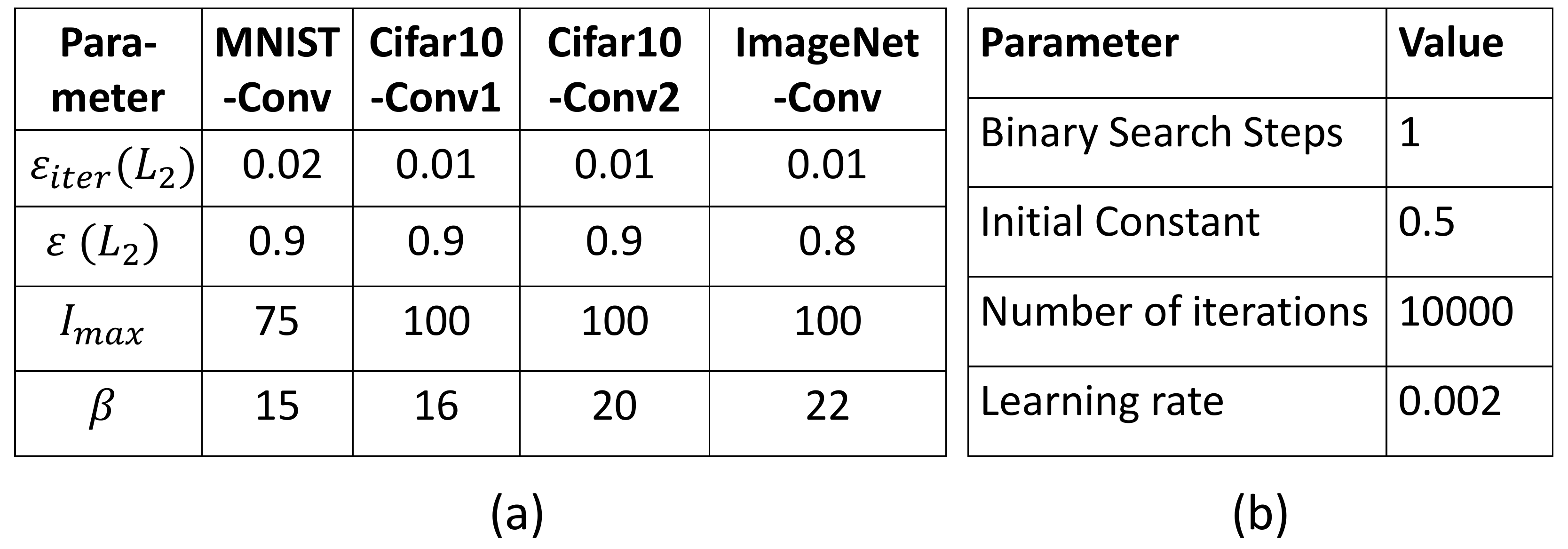}
 \caption{Hyper-parameter values used for a) White-Box sparsity attack and b) Stage 2 of the Black-Box sparsity attack}
 \label{table:attacks}
 \vspace{-5pt}
\end{figure}

The attack framework is implemented using TensorFlow \cite{TF}. We utilize the software framework provided in \cite{Zoo} to realize the second stage of the black-box attack. The different hyper-parameters used by the white-box and black-box sparsity attacks are listed in Figures~\ref{table:attacks}(a) and \ref{table:attacks}(b), respectively. Additionally, for the white-box attack, we set $O_{max}$ to 1, and $c_{in}$, $c_{min}$ and $c_{max}$ to 0.5, 0 and 1, respectively, in all our experiments. The impact of these hyper-parameters is discussed in Section~\ref{subsec:whiteboxresult}. The first stage of the black-box attack is conducted using the same parameters as listed in Figure~\ref{table:attacks}(a). The substitute models used in this step are discussed in Section~\ref{subsec:blackboxresult}.

\begin{figure}[h!]
 \includegraphics[width=0.98\columnwidth, scale=1]{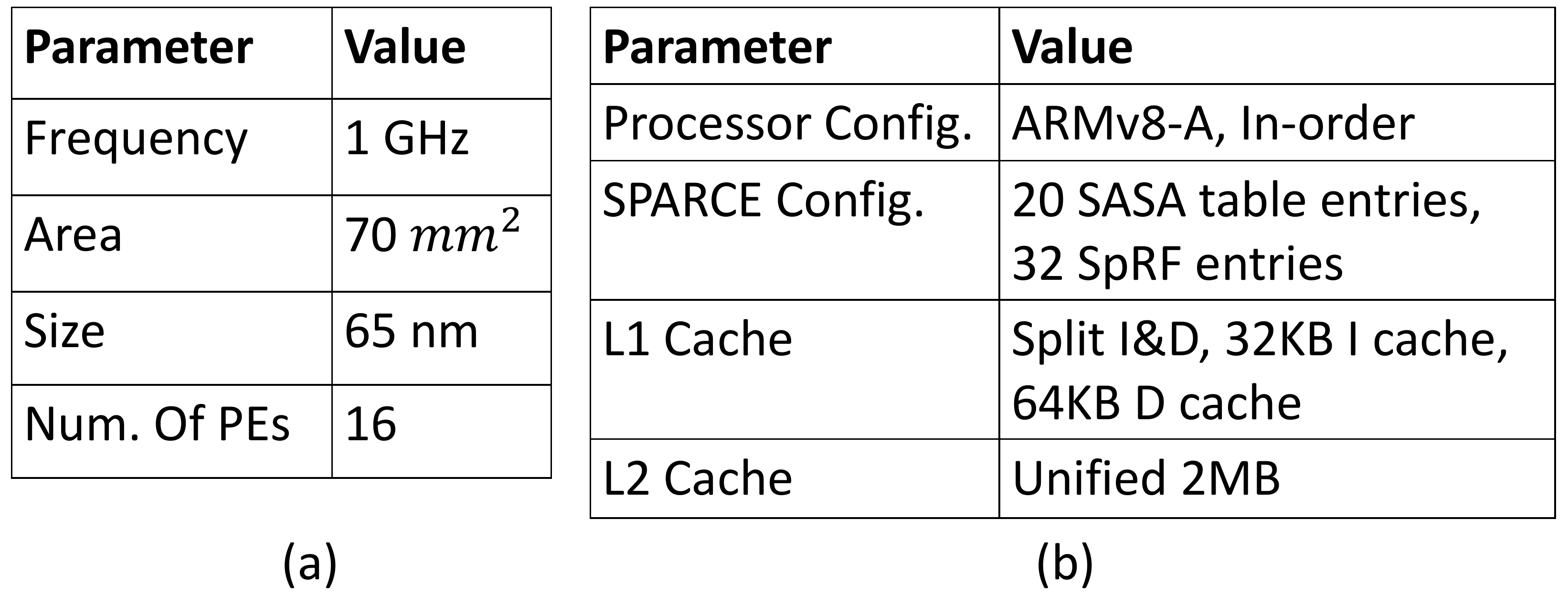}
 \caption{Micro-architectural details for a) Cnvlutin and b) S\textsc{par}CE}
 \label{table:arch}
 \vspace{-7pt}
\end{figure}

\textbf{Sparsity-optimized platforms.} We evaluated the effects of adversarial sparsity attacks on two different sparsity-optimized platforms, namely, the Cnvlutin DNN accelerator \cite{Cnvlutin} and the S\textsc{par}CE general-purpose processor \cite{Sparce}. The micro-architectural details of each of these platforms are listed in Figure~\ref{table:arch}. For measuring execution times on Cnvlutin, we develop a cycle-accurate simulator using the details provided in \mbox{\cite{Cnvlutin}}. Our  simulator closely matches the execution time and energy consumption results reported in \mbox{\cite{Cnvlutin}}. We utilize a simulator provided to us by the authors of S\textsc{par}CE for measuring execution time and energy consumption. 

\vspace{-2pt}
\section{Results}
\label{sec:expResults}
In this section, we present experimental results and evaluate the effectiveness of the proposed adversarial sparsity attacks.
\vspace{-5pt} 
\subsection{White-Box Sparsity Attacks}
\label{subsec:whiteboxresult}
\subsubsection{Impact on Activation Sparsity, Execution time and EDP} 
\vspace{2pt}
We first evaluate the decrease in activation sparsity, and corresponding increases in execution time and energy-delay product, resulting from white-box adversarial sparsity attacks.

\begin{figure}[h!]
\centering
\vspace*{-5pt}
 \includegraphics[width=\columnwidth]{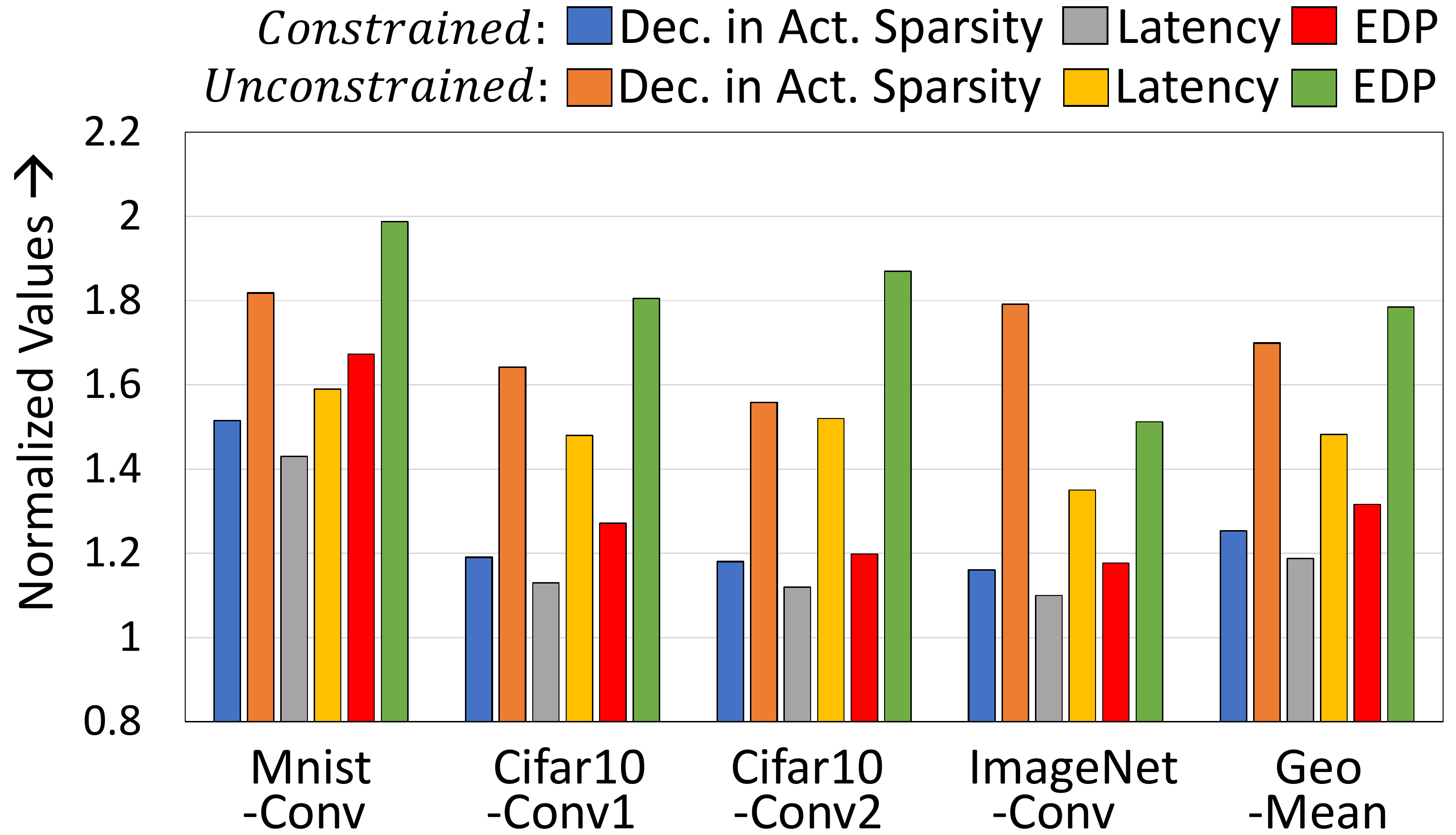}
 \caption{Impact of white-box adversarial sparsity attacks on activation sparsity, execution time and EDP on Cnvlutin}
 \label{fig:Cnvlutin}
 \vspace{-7pt}
\end{figure}

\textbf{Decrease in Activation Sparsity}: The decreases in activation sparsity achieved by the white-box sparsity attack under two different input perturbation constraints are shown in Figure~\ref{fig:Cnvlutin}. In certain scenarios where human supervision of both the input and output of the DNN is present, it is necessary that the sparsity attack meets all criteria listed in Section~{\ref{sec:design}}. Specifically, the added input distortions should be imperceptible, \emph{i.e.}, constrained within the $L_{2}$ bound specified in Figure 9. We refer to these attack scenarios as constrained. However, in scenarios such as autonomous self-driving where there is no human supervision of the input, this criterion can be relaxed, \emph{i.e.}, the input perturbations can be of any magnitude provided the inputs are within a valid range. The only behavior observable to the user in such scenarios is the correctness of the classification. Such an attack scenario is referred to as unconstrained. Across the benchmark DNNs, the proposed white-box sparsity attack achieves a
1.16$\times$-1.52$\times$ (average: 1.25$\times$) decrease in activation sparsity for the constrained case. These values increase to 1.55$\times$-1.82$\times$ (average: 1.70$\times$) for the unconstrained scenario.

As the activation sparsity of the networks considered vary from roughly 50\%-70\%, a trivial limit for the maximum reduction in sparsity possible would range from 2$\times$-3$\times$. A more meaningful and tight limit is difficult to establish because at the heart of the attack lies a non-convex optimization problem with upto tens of thousands of variables (corresponding to the input size).

The reduction in sparsity achieved by the attack translates to varying degrees of execution time and energy increase depending on the DNN, as discussed next.


\textbf{Increase in Execution Time:} Figure~\ref{fig:Cnvlutin} also shows the impact of adversarial sparsity attacks on the execution time of different networks on the Cnvlutin~\cite{Cnvlutin} accelerator. We observe a slowdown of 1.12$\times$-1.43$\times$ (average: 1.19$\times$) across all networks for the constrained scenario, which increases to 1.37$\times$-1.59$\times$ (average: 1.48$\times$) for the unconstrained case.  
Comparing results across datasets, we note that for some datasets, sparsity may be present in the input image itself, allowing the sparsity attack to also target the first layer of the network in addition to the subsequent layers. Specifically, the images in the MNIST dataset are sparse ($\sim$80\%), while the CIFAR-10 and ImageNet datasets exhibit minimal (0-0.3\%) sparsity in the input. Thus, for the CIFAR-10 and ImageNet networks, decreases in sparsity and the corresponding impact on execution time and energy are possible only from the second convolutional layer onwards. As a result, networks with comparatively smaller runtimes for the first convolutional layer, such as Cifar10-Conv2 and ImageNet-Conv feel the impact of decreased sparsity in a more pronounced fashion unlike networks such as Cifar10-Conv1, whose first convolutional layer alone nearly takes up 26\% of the total runtime. We note that the attack is likely to have an even stronger impact on more recent sparse accelerators \cite{sparten, SCNN}, as they show greater sensitivity to sparsity. 


\textbf{Increase in EDP:} We utilize Energy-Delay Product (EDP) as a metric to evaluate the effect of adversarial sparsity attacks on energy efficiency of DNN execution. 
Sparsity attacks cause an increase in energy consumption due to the higher number of operations to be performed. This results in a 1.18$\times$-1.67$\times$ (average: 1.32$\times$) 
increase in EDP for Cnvlutin 
when the perturbations are constrained. When the constraints are relaxed, the increase in EDP ranges from 1.51$\times$-1.99$\times$ (average: 1.79$\times$). 

\vspace{7pt}
\subsubsection{Impact on the activation sparsity distribution}
We observe 
\begin{wrapfigure}{r}{0.66\columnwidth}
\centering
 \includegraphics[width=0.66\columnwidth]{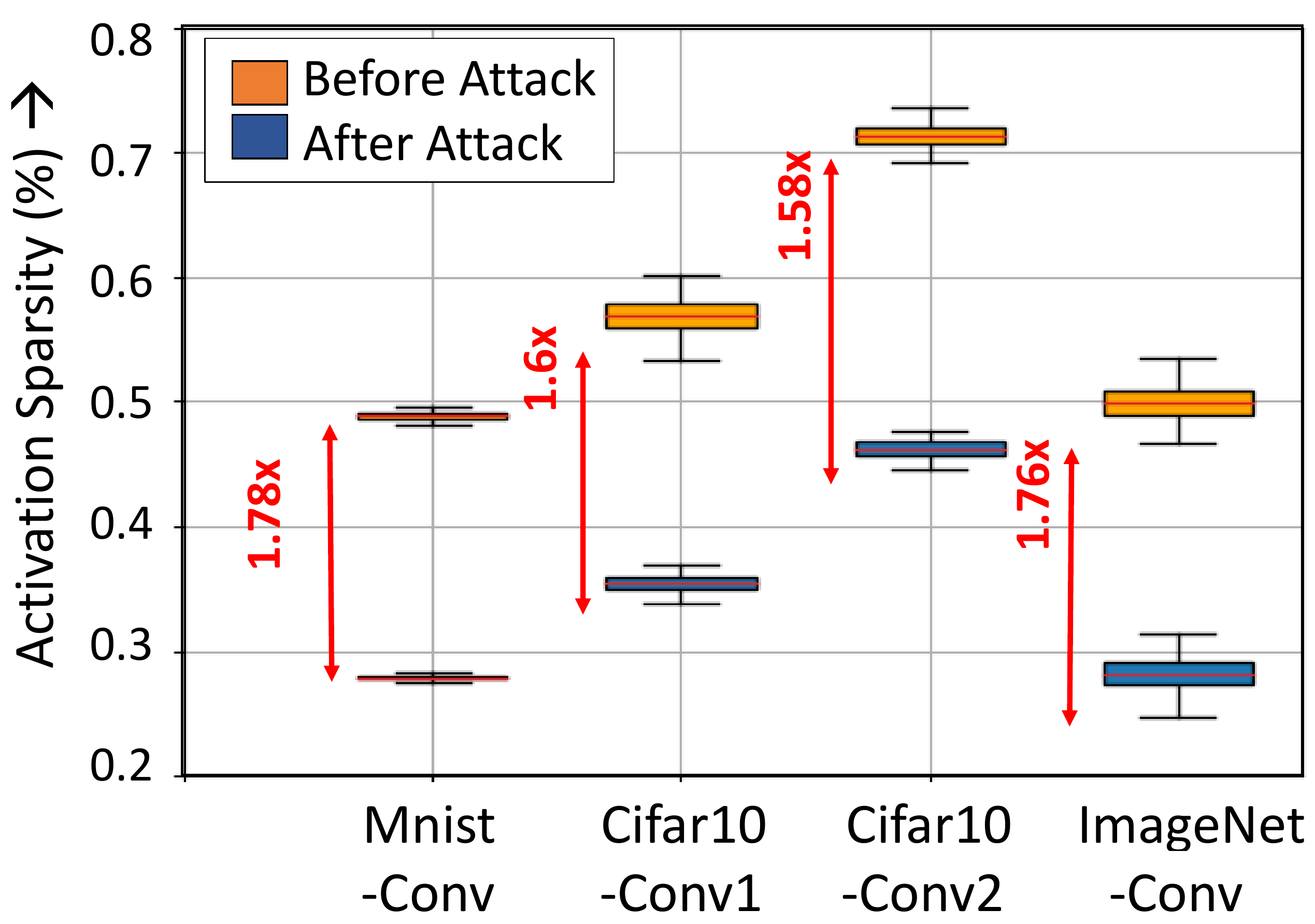}
 \caption{Distribution of activation sparsity before and after application of the attack}
 \label{fig:dist_act_sparse}
 \vspace{-4pt}
\end{wrapfigure}the change in the activation sparsity distribution across the 
test dataset images as a result of the sparsity attack. Figure~{\ref{fig:dist_act_sparse}} depicts a box-whisker plot of the activation sparsity distribution on all networks, before and after the application of the sparsity attack. As marked by the red arrows, the worst case sparsity of the adversarial inputs is 1.58$\times$-1.78$\times$ lower than the worst case sparsity of the clean inputs. Clearly, even if designers account for the worst case sparsity present in the clean inputs, the attack reduces sparsity further by a significant degree, causing inference time and energy to exceed the limits of even a conservative design.

\begin{wrapfigure}{l}{0.71\columnwidth}
\centering
 \includegraphics[width=0.71\columnwidth]{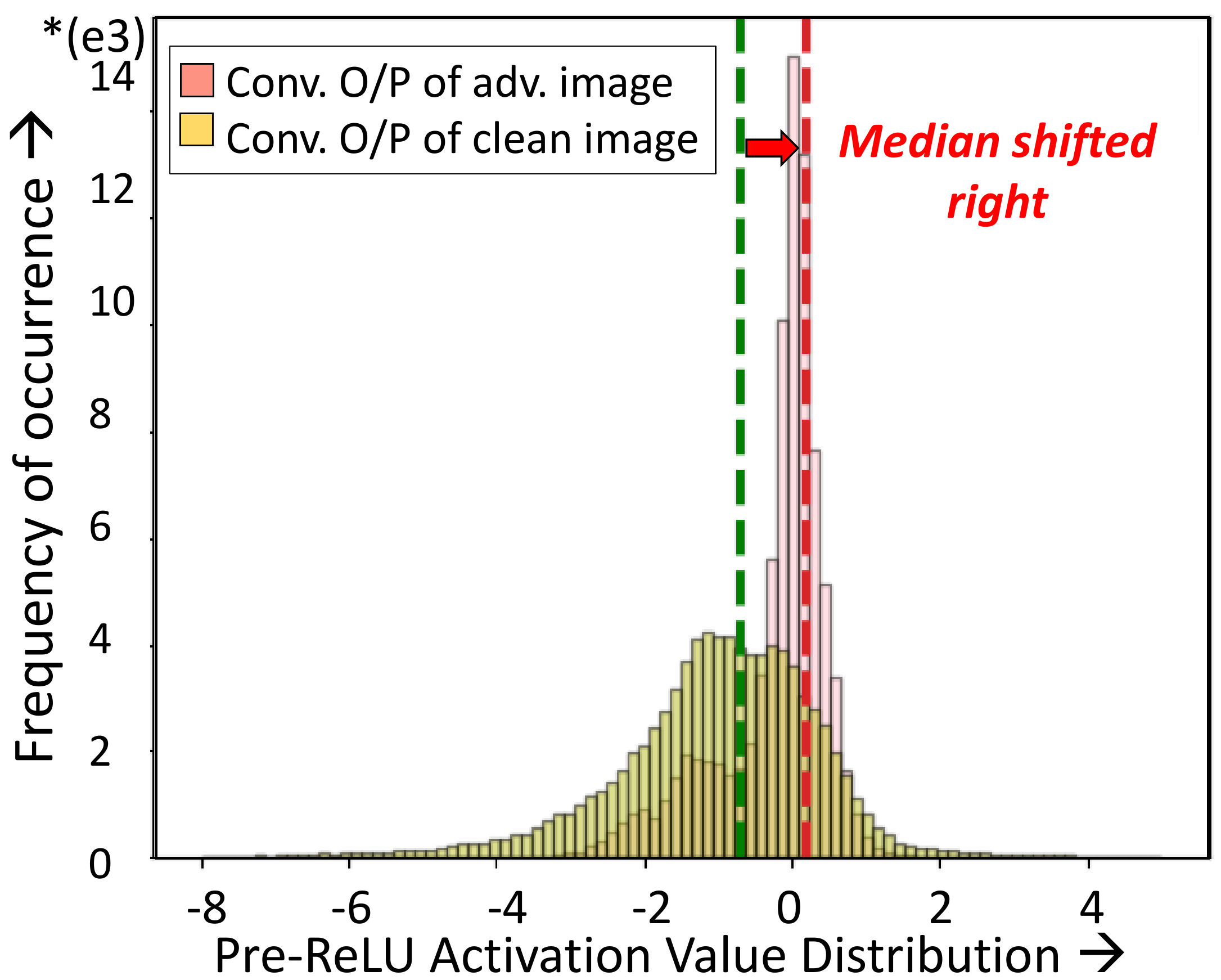}
 \caption{Distribution of activation values at the output of the first convolutional layer of Cifar10-Conv2}
 \label{fig:dist_act_val}
\end{wrapfigure}

Additionally, we also provide a visualization of the distribution of values at the output of a convolutional layer prior to the application of the ReLU activation. In Figure~{\ref{fig:dist_act_val}} we depict the 
pre-ReLU activations at the output of the first convolutional layer of Cifar10-Conv2 before and after the attack has been launched. The figure indicates that the negative pre-ReLU activatons are shifted towards the right under the adversarial input, i.e., the activation value distribution becomes more positive. This leads to a decrease in post-ReLU activation sparsity. However, note that the pre-ReLU activations are limited to values of small magnitude, which helps in preserving the network output, and hence, accuracy.

\vspace{5pt}
\subsubsection{Analyzing impact of different hyper-parameters}\text{ }\text{ }\text{ }\text{ }\text{ }\text{}

\textbf{Impact of $c$:} We now analyze the performance of Algorithm~\ref{alg:ASA_im} in terms of the  hyper-parameter $c$ and study the trade-off between decrease in activation sparsity and loss in classification accuracy. When $c =0$, the attack solely focuses on reducing 
\begin{wrapfigure}{r}{0.6\columnwidth}
\centering
 \includegraphics[width=0.6\columnwidth,scale=1]{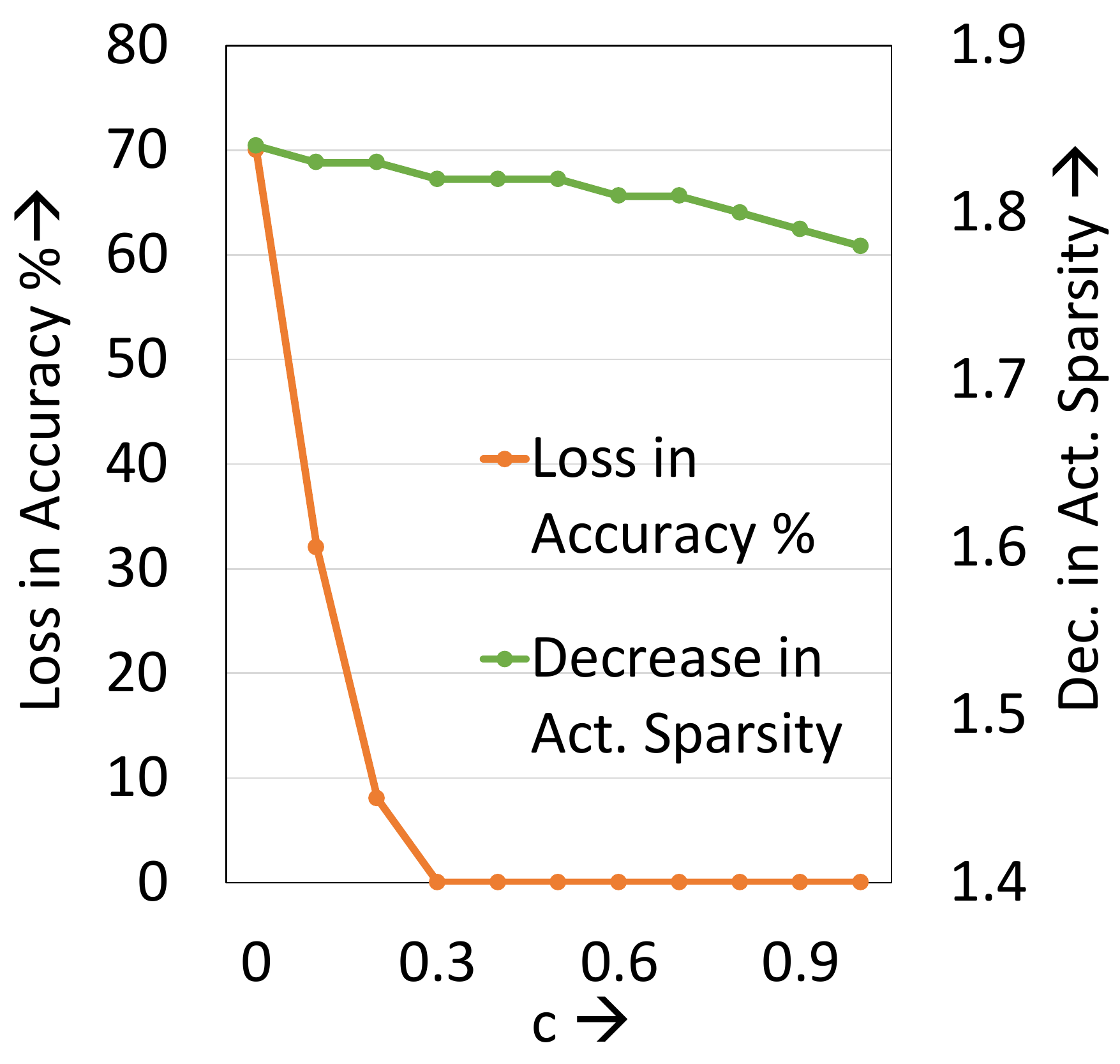}
 \caption{ Impact of $c$ on performance of Algorithm~\ref{alg:ASA_im}}
 \label{fig:VarC}
 \vspace*{-5pt}
\end{wrapfigure}
the activation sparsity of the network, and does not factor classification performance. On Mnist-Conv, this point translates to nearly 1.85$\times$ decrease in activation sparsity on average across images, but a 70\% decrease in classification performance. Interestingly, Algorithm~\ref{alg:ASA_im} identifies a value of $c = 0.3$, at which there is absolutely \textit{no loss in classification accuracy} and 1.83$\times$ decrease in activation sparsity, which is virtually the same as thee activation sparsity obtained at $c = 0$. These results indicate the ability of the algorithm to identify points on the objective landscape that ensure virtually no impact on classification accuracy, while only marginally compromising on the decrease in activation sparsity. 

\begin{figure}[h!] 
\centering
 \includegraphics[width=\columnwidth,scale=1]{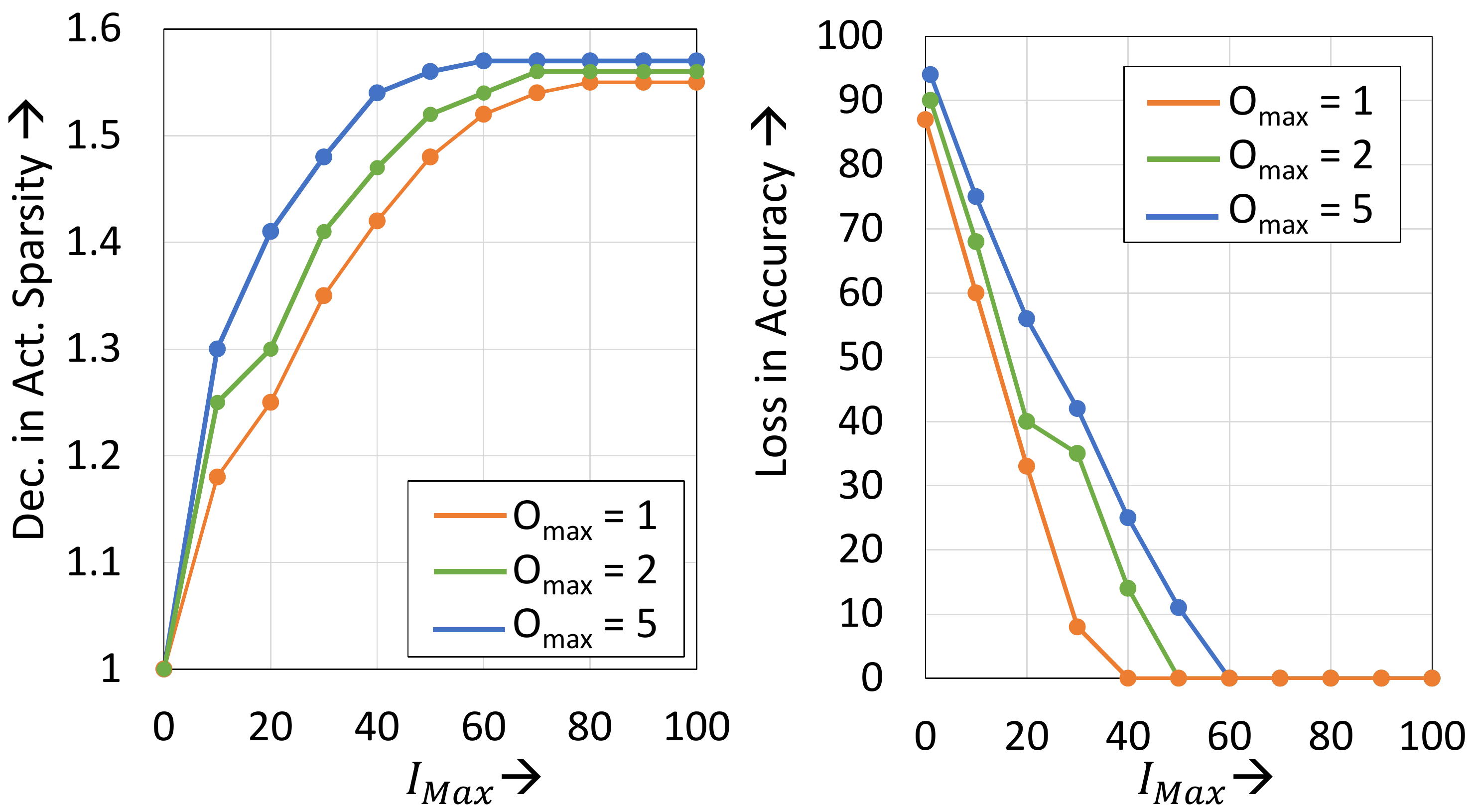}
 \caption{ Impact of $O_{max}$ and $I_{max}$ on performance of Algorithm~\ref{alg:ASA_im}}
 \label{fig:VarIter}
\end{figure}

\textbf{Impact of $O_{max}$ and $I_{max}$:} In Figure~\ref{fig:VarC}, the presence of a plateau-region in both the sparsity and accuracy curves for a wide range of $c$ values indicates the low sensitivity of the performance of Algorithm~\ref{alg:ASA_im} to $c$ in this range - an insight that we exploit to drastically reduce the number of binary-search iterations used to find an optimal $c$. Across our networks, providing any $c_{in}$ in the range of 0.5 to 1 achieves adequate results in merely 1 outer-loop iteration. This is further underscored in Figure~\ref{fig:VarIter}, which depicts the minimal impact of increasing $O_{max}$ on the performance of Algorithm~\ref{alg:ASA_im} when $c_{min}$= 0.2, $c_{in}$= 0.6 and $c_{max}$= 1, in the context of Cifar10-Conv2. 
Across all curves, for each outer-loop iteration it typically takes around 40-60 inner-loop iterations for the accuracy to converge, and an additional 20-40 iterations for the decrease in activation sparsity to saturate. All values of $O_{max}$ considered (from 1 to 5) resulted in correct predictions after Algorithm~\ref{alg:ASA_im} terminated. More importantly, we observe that higher values of $O_{max}$, which expend more effort towards finding a better $c$  provide only 2-3\% reduction in activation sparsity over $O_{max} = 1$. Therefore, across our networks, we conduct our white-box attacks with $O_{max}$ set to 1, ensuring a runtime-efficient attack with little to no sacrifice in efficacy.

\subsubsection{Runtime Analysis} \label{sec:Runtime}
\begin{figure}[h] 
\centering
 \includegraphics[width=\columnwidth,scale=1]{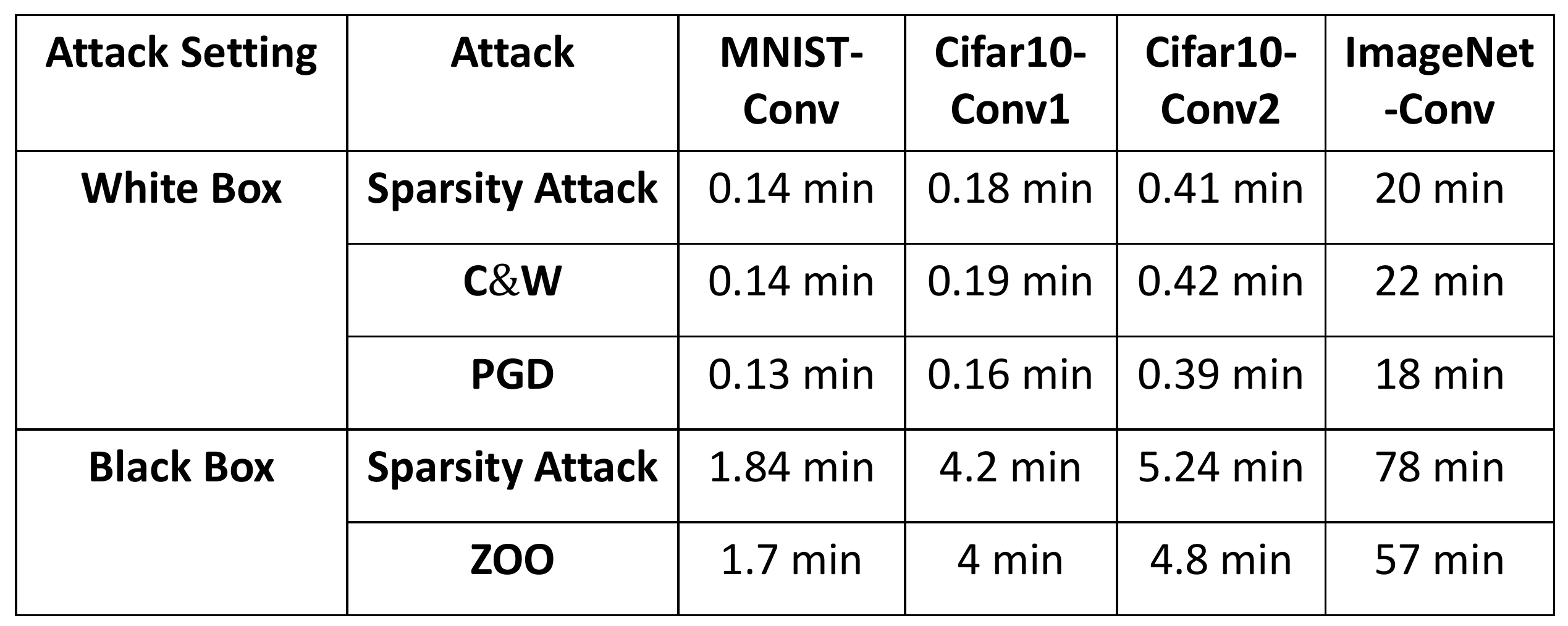}
 \caption{Runtimes per image for sparsity attacks and conventional accuracy-based attacks}
 \label{fig:VarRuntime}
\vspace*{-0.2in}
\end{figure}

We list the runtimes involved in launching the white-box sparsity attack on each network in Figure~\ref{fig:VarRuntime}, with all experiments conducted on a single NVIDIA RTX 2080Ti GPU. For reference, we also compare our runtimes against state-of-the-art accuracy-based attacks namely Projected Gradient Descent (PGD) \cite{Madry} and Carlini and Wagner(C$\&$W) \cite{CW} on the same networks, for an equal number of total iterations. As observed, the runtime costs of our attack are on average only marginally higher than PGD (by 11\%), and faster than C$\&$W (by 6\%). This highlights the computational feasibility of the proposed adversarial sparsity attacks. 

\vspace*{-5pt}
\subsection{Black-box attacks} \label{subsec:blackboxresult}
\subsubsection{Impact on Activation Sparsity, Execution Time and EDP}
We study the effect of black-box sparsity attacks on all networks for the unconstrained distortion scenario, and demonstrate the increase in execution time and EDP on the Cnvlutin accelerator. For the Cifar10 and MNIST networks, we use 
\begin{wrapfigure}{r}{0.75\columnwidth}
\centering
 \includegraphics[width=0.75\columnwidth]{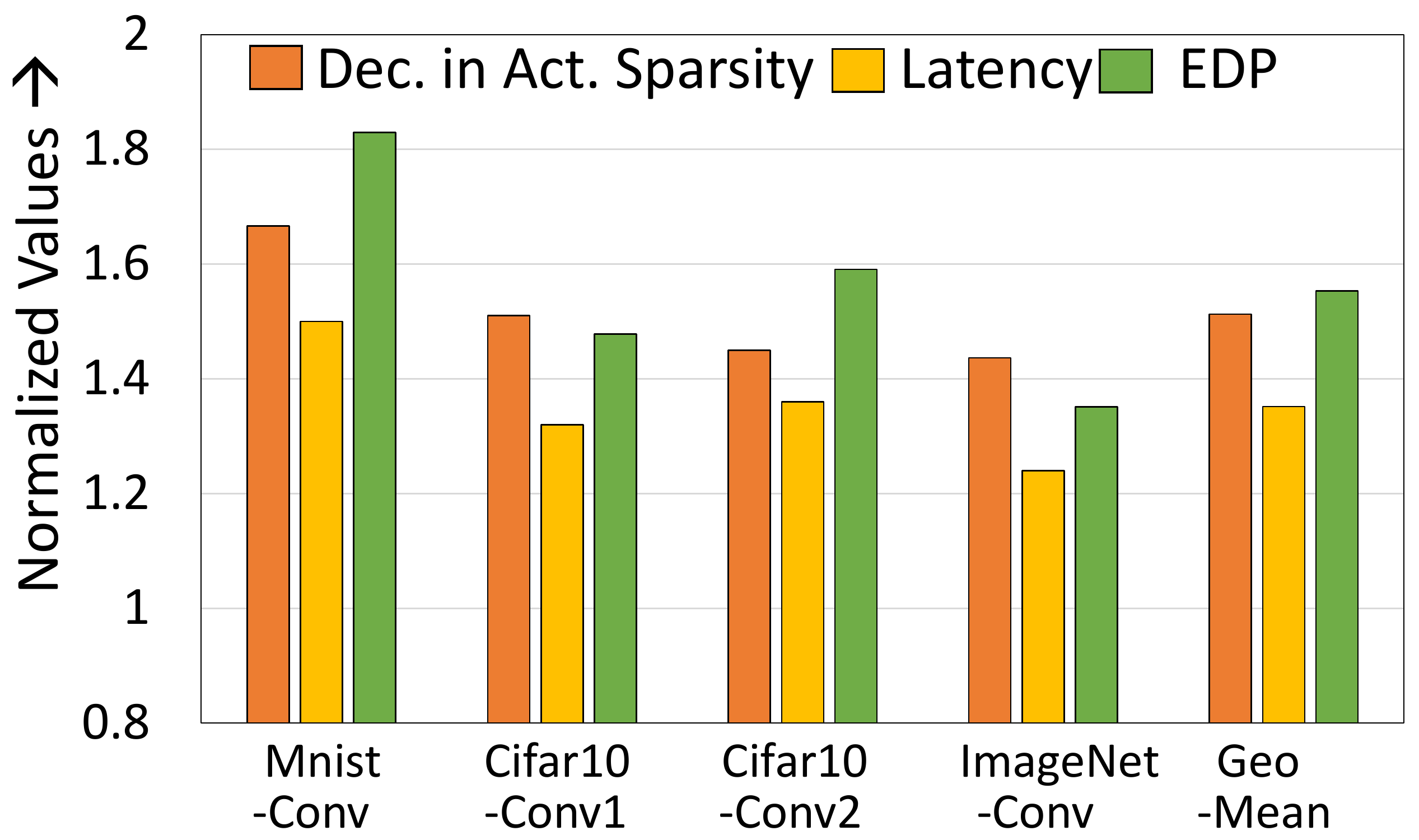}
 \caption{Impact on Activation Sparsity, Latency and EDP for Black-Box attacks}
 \label{fig:BB}
\end{wrapfigure}
Model 1 described in Figure~\ref{fig:VarBB_tab} as the substitute model, and employ the VGG-19 \cite{VGG19} architecture as the substitute when attacking the ImageNet network. The white-box sparsity attack causes a 1.53$\times$ and 1.48$\times$ sparsity reduction on Model 1 for the Cifar10 and MNIST datasets respectively, and a 1.65$\times$ reduction in sparsity for the VGG-19 model. As shown in Figure~\ref{fig:BB}, we achieve a 1.43$\times$ - 1.67$\times$ decrease in activation sparsity, which translates to 1.24$\times$-1.5$\times$ increase in execution time, and 1.35$\times$-1.62$\times$ increase in EDP. These results demonstrate that sparsity attacks can be effectively launched in black-box scenarios as well. 

\vspace{9pt}
\subsubsection{Impact of Substitute Model on Performance}
\begin{wrapfigure}{r}{0.6\columnwidth}
\centering
 \includegraphics[width=0.6\columnwidth,scale=1]{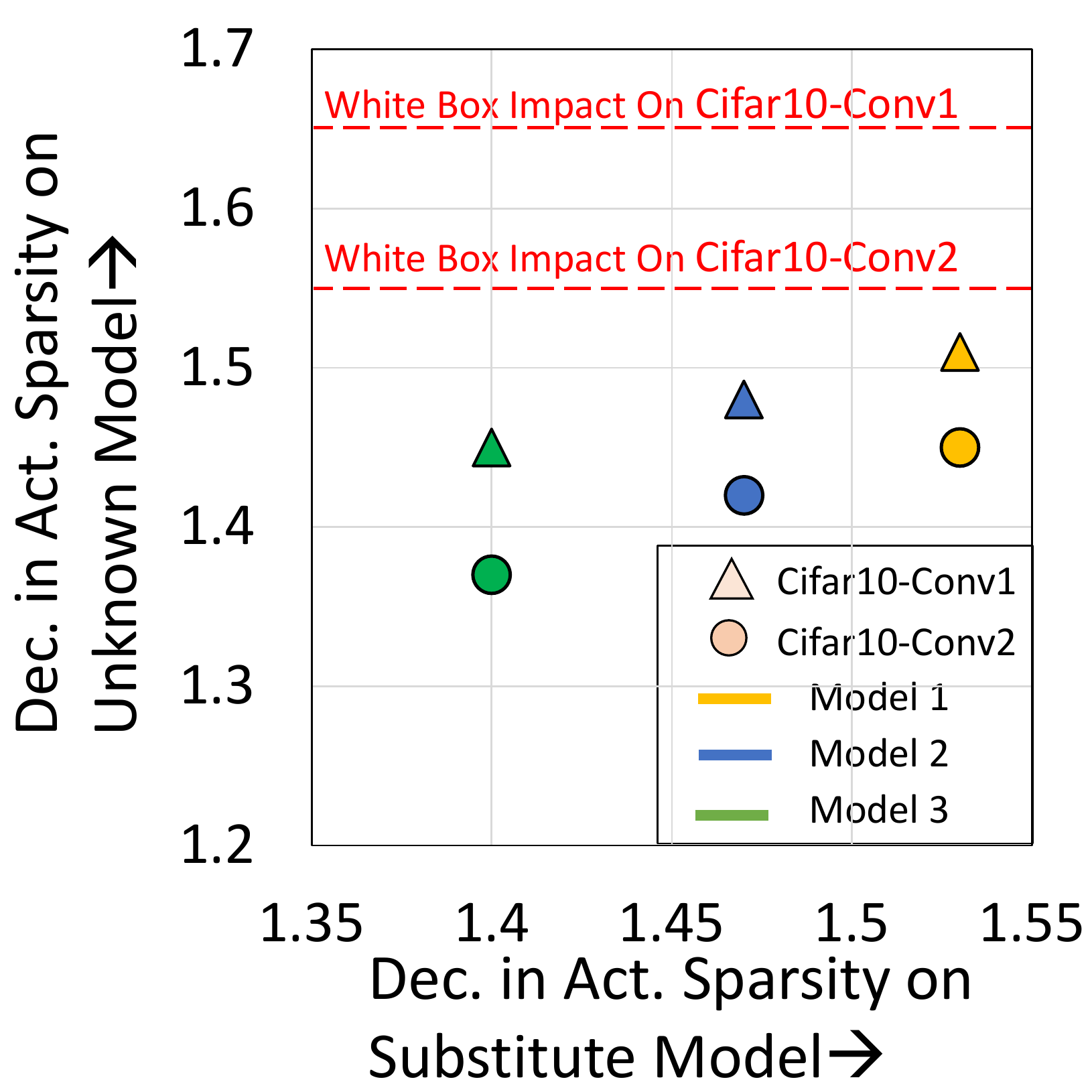}
 \caption{ Analyzing impact of Substitute Model}
 \label{fig:VarSubM}
\vspace*{-3pt}
\end{wrapfigure}

Figure~\ref{fig:VarSubM} illustrates the impact of using different substitute models when launching a black-box 
attack on Cifar10-Conv1 and Cifar10-Conv2, with the models listed in Figure~\ref{fig:VarBB_tab} used as substitute models. The effectiveness of the transfer is dictated by two factors: a) the inherent vulnerability of the unknown network to the sparsity attack and b) the vulnerability of the substitute model deployed. Our experiments suggest that, the final reduction in sparsity achieved by the black-box attack is generally upper bounded by impact of the white-box attack on the same DNN. Moreover, in certain cases, if the unknown model is highly susceptible to the attack, it can exhibit a higher reduction in sparsity than that incurred by the substitute model. This trend is observed when using Model 3 to attack Cifar10-Conv1. Further, amongst the different substitute models deployed, we empirically find that substitute models that exhibit higher vulnerability to the white-box attack are more likely to induce a transfer of higher potency.

\vspace{7pt}
\subsubsection{Runtime Analysis} \label{sec:Runtime}
Figure~\ref{fig:VarRuntime} also lists the runtimes of black-box attacks across different networks. The higher runtime of the black-box sparsity attack compared to the white-box version is attributed to the second stage, which employs a targeted accuracy-based attack to ensure that the perturbed input does not cause a change in the DNN's output. Such targeted black-box attacks have been reported to be significantly more expensive to compute than their white-box counterparts. This is due to the increased difficulty in estimating the gradients that are critical towards attaining high success rates. For example, conducting the black-box sparsity attack on Cifar10-Conv1 for example takes 4.2 min per image, which is far higher than the 0.18 min per image required for the white-box sparsity attack. Comparing the runtimes of black-box sparsity attacks against that of the ZOO attacks on the same networks, we find that our attack is on average merely 13\% more expensive to compute, due to the relatively small overhead of the first stage.

\vspace{-7pt}
\subsection{Analysis of defense techniques}
\begin{figure}[h]
\centering
\vspace*{-5pt}
 \includegraphics[width=\columnwidth]{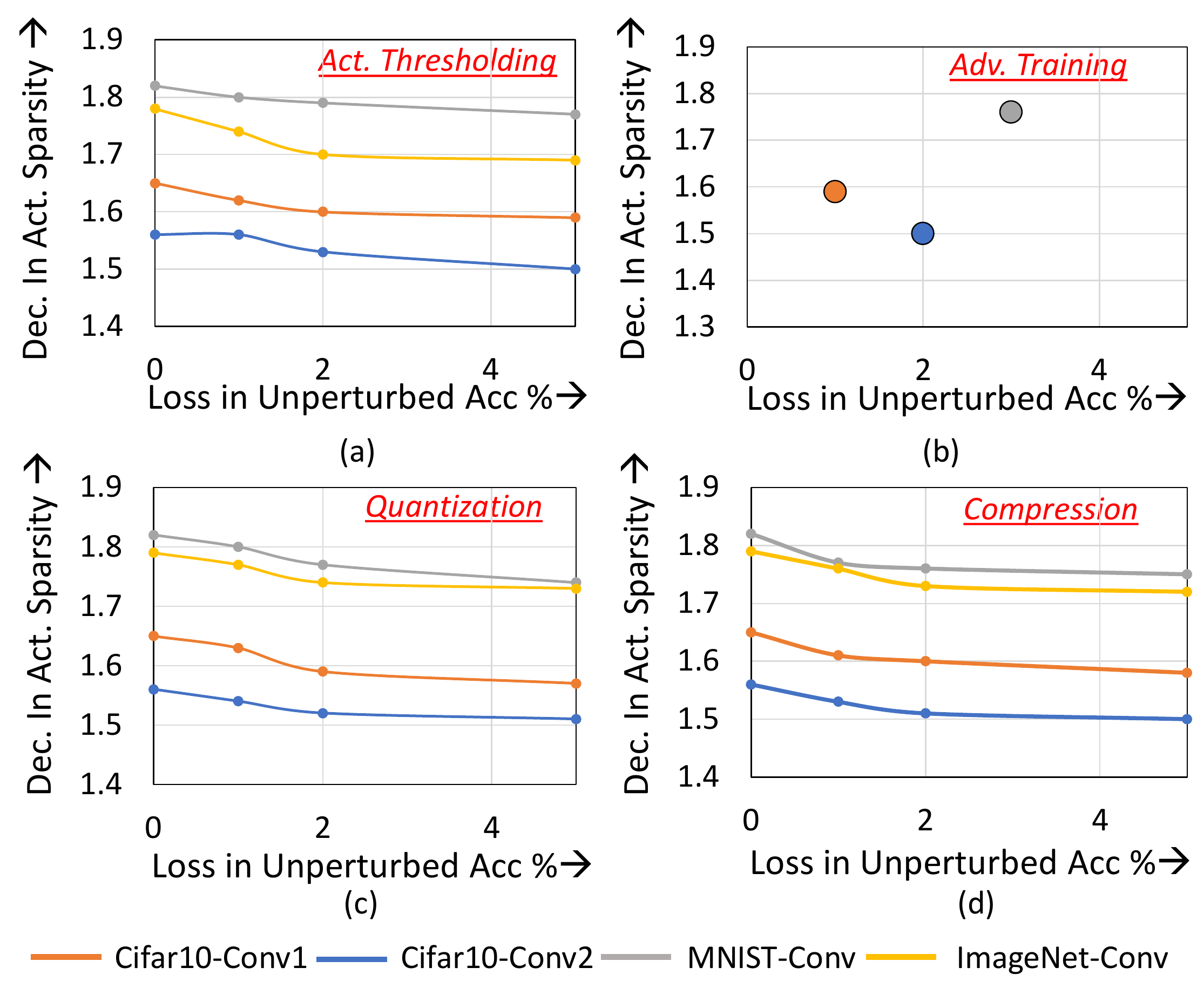}
 \caption{Analyzing the impact of different defense techniques}
 \label{fig:VarDef}
\vspace*{-5pt}
\end{figure}

In this subsection, we study the effectiveness of the defense techniques discussed in Section~\ref{sec:defense} against the white-box attacks, in the unconstrained distortion scenario. Figure~\ref{fig:VarDef}(a) illustrates the trade-off between the loss in unperturbed accuracy and the decrease in activation sparsity for the activation thresholding defense. We report results for four configurations that attempt to restore sparsity while incurring within 0\%, 1\%, 2\% and 5\% drop in unperturbed accuracy, respectively. The configurations that provide no loss in accuracy have negligible effect on the impact of the attack. Further, even with a $5$\% loss in accuracy, the defense is only able to degrade the attack's impact on sparsity by 6-8\% across networks. 

The results for input quantization [Figure~\ref{fig:VarDef}(c)] and compression [Figure~\ref{fig:VarDef}(d)] tell a similar story. On Cifar10-Conv1 an input bit-width of 4, or a compression quality of 85 (out of 100) mitigates the attack's impact by 5-6\%, while incurring a 5\% drop in unperturbed accuracy. It is evident that, to mitigate the impact of sparsity attacks considerably, the defender must resort to using extremely low bitwidths and compression qualities for the input, which generally cause a severe reduction in unperturbed accuracy. We report similar findings for adversarial training, as seen in Figure~\ref{fig:VarDef}(b).
\subsection{Evaluation on general-purpose processors}
\begin{figure}[h!]
\centering
 \includegraphics[width=\columnwidth, scale = 0.95]{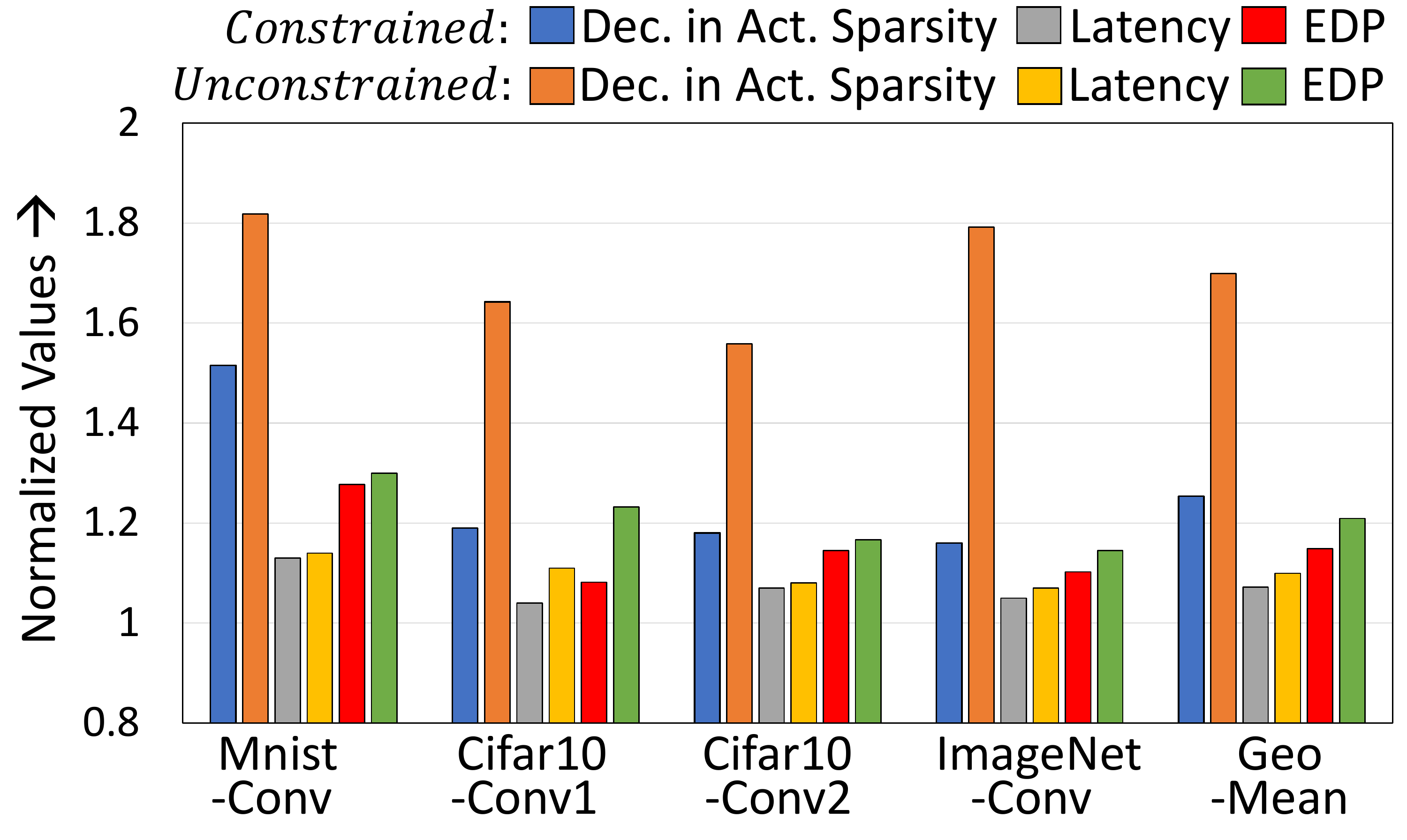}
 \caption{Impact of adversarial sparsity attacks on activation sparsity, execution time and EDP on S\textsc{par}CE}
 \label{fig:Sparce}
\end{figure} 
\vspace{-5pt}
In this subsection, we analyze the impact of the sparsity attack on a sparsity-optimized general purpose processor, SparCE. For the sake of brevity, we consider the impact of the white-box attack alone, and depict the resulting increases in latency and EDP in Figure~\ref{fig:Sparce}. The increase in latency is 1.04$\times$-1.14$\times$, and the increase in EDP is 1.08$\times$-1.30$\times$ - clearly lower than the increase in latency and EDP affected on the Cnvlutin accelerator. We attribute this to the fact that general-purpose processor based platforms derive lower benefits from sparsity in the first place due to overheads such as control instructions for pointer arithmetic, loop counters, {\em etc.} Thus, it is not surprising that they tend to show lower vulnerability to the sparsity attacks compared to hardware accelerators.


\section{Conclusion}
\label{sec:conclusion}
We present a new class of adversarial attacks that adversely impact the execution time and energy consumption of DNNs on sparsity-optimized platforms by introducing input perturbations that cause a reduction in activation sparsity. The proposed sparsity attacks differ from conventional accuracy-based attacks as they do not affect classification accuracy, and can be launched in both white and black-box settings. Across our suite of 4 DNNs on 3 datasets, we achieve a 1.16$\times$-1.82$\times$ decrease in activation sparsity for no loss in classification accuracy. We also demonstrate the impact of our attack on a sparsity-optimized DNN accelerator, achieving a 1.12$\times$-1.59$\times$ increase in latency and a 1.18$\times$-1.99$\times$ increase in EDP.
\bibliographystyle{unsrt}
\bibliography{main}

\begin{thebibliography}{10}

\bibitem{DRL}
K.~He et. al.
\newblock Deep residual learning for image recognition.
\newblock {\em CoRR}, abs/1512.03385, 2015.

\bibitem{DeepSpeech}
Awni Y.~Hannun et~al.
\newblock Deep speech: Scaling up end-to-end speech recognition.
\newblock {\em CoRR}, abs/1412.5567, 2014.

\bibitem{Hinton}
G.~{Hinton} et~al.
\newblock Deep neural networks for acoustic modeling in speech recognition: The
  shared views of four research groups.
\newblock {\em IEEE Signal Processing Magazine}, 29(6):82--97, Nov 2012.

\bibitem{DeepCompression}
Song~Han et~al.
\newblock Deep compression: Compressing deep neural networks with pruning,
  trained quantization and huffman coding.
\newblock 10 2016.

\bibitem{SCNN}
A.~Parashar et~al.
\newblock {SCNN: An Accelerator for Compressed-sparse Convolutional Neural
  Networks}.
\newblock {\em SIGARCH Comput. Archit. News}, 45(2):27--40, June 2017.

\bibitem{Cnvlutin}
J.~{Albericio} et~al.
\newblock Cnvlutin: Ineffectual-neuron-free deep neural network computing.
\newblock In {\em ISCA 2016}, pages 1--13, June 2016.

\bibitem{eyerissv2}
Y.~{Chen}, T.~{Yang}, J.~{Emer}, and V.~{Sze}.
\newblock Eyeriss v2: A flexible accelerator for emerging deep neural networks
  on mobile devices.
\newblock {\em IEEE Journal on Emerging and Selected Topics in Circuits and
  Systems}, 9(2):292--308, 2019.

\bibitem{Sparce}
S.~{Sen} et~al.
\newblock {SparCE: Sparsity Aware General-Purpose Core Extensions to Accelerate
  Deep Neural Networks}.
\newblock {\em IEEE Transactions on Computers}, 68(6):912--925, June 2019.

\bibitem{sparten}
A.~Gondimalla et~al.
\newblock Sparten: A sparse tensor accelerator for convolutional neural
  networks.
\newblock In {\em Proc. MICRO}, page 151–165, 2019.

\bibitem{extensor}
K.~Hegde et~al.
\newblock Extensor: An accelerator for sparse tensor algebra.
\newblock In {\em Proc. MICRO}, page 319–333, 2019.

\bibitem{sparseTC}
Maohua Zhu~et al.
\newblock Sparse tensor core: Algorithm and hardware co-design for vector-wise
  sparse neural networks on modern gpus.
\newblock In {\em Proc. MICRO}, page 359–371, 2019.

\bibitem{VGG19}
Karen Simonyan and Andrew Zisserman.
\newblock Very deep convolutional networks for large-scale image recognition.
\newblock {\em CoRR}, abs/1409.1556, 2014.

\bibitem{self_driving_car}
Self driving car reaction time.
\newblock http://news.mit.edu/2019/how-fast-humans-react-car-hazards-0807.

\bibitem{cifar10}
Alex Krizhevsky, Vinod Nair, and Geoffrey Hinton.
\newblock Cifar-10 (canadian institute for advanced research).

\bibitem{IFGSM}
Alexey Kurakin, Ian~J. Goodfellow, and Samy Bengio.
\newblock Adversarial examples in the physical world.
\newblock {\em CoRR}, abs/1607.02533, 2016.

\bibitem{deepface}
Yaniv Taigman~et al.
\newblock Deepface: Closing the gap to human-level performance in face
  verification.
\newblock In {\em Proceedings of the 2014 IEEE Conference on Computer Vision
  and Pattern Recognition}, CVPR ’14, page 1701–1708, USA, 2014. IEEE
  Computer Society.

\bibitem{gpt3}
Tom B.~Brown et~al.
\newblock Language models are few-shot learners, 2020.

\bibitem{CW}
N.~{Carlini} et~al.
\newblock Towards evaluating the robustness of neural networks.
\newblock In {\em 2017 IEEE Symposium on Security and Privacy (SP)}, pages
  39--57, May 2017.

\bibitem{MIM}
Y.~Dong et~al.
\newblock Discovering adversarial examples with momentum.
\newblock {\em CoRR}, abs/1710.06081, 2017.

\bibitem{Zoo}
Pin-Yu~Chen et~al.
\newblock {ZOO}.
\newblock {\em Proceedings of the 10th ACM Workshop on Artificial Intelligence
  and Security - AISec ’17}, 2017.

\bibitem{SimpleBB}
C.~Guo et~al.
\newblock Simple black-box adversarial attacks.
\newblock {\em CoRR}, abs/1905.07121, 2019.

\bibitem{BB}
Nicolas~Papernot et~al.
\newblock Practical black-box attacks against deep learning systems using
  adversarial examples.
\newblock {\em CoRR}, abs/1602.02697, 2016.

\bibitem{tsc}
V.~Duddu et~al.
\newblock Stealing neural networks via timing side channels.
\newblock {\em CoRR}, abs/1812.11720, 2018.

\bibitem{membership_inference}
Reza~Shokri et~al.
\newblock Membership inference attacks against machine learning models.
\newblock {\em CoRR}, abs/1610.05820, 2016.

\bibitem{bitflip}
A.~S.~Rakin et~al.
\newblock Bit-flip attack: Crushing neural network withprogressive bit search.
\newblock {\em CoRR}, abs/1903.12269, 2019.

\bibitem{survey}
M.~Isakov et~al.
\newblock Survey of attacks and defenses on edge-deployed neural networks,
  2019.

\bibitem{fault_inj}
Yannan Liu, Lingxiao Wei, Bo~Luo, and Qiang Xu.
\newblock Fault injection attack on deep neural network.
\newblock In {\em Proceedings of the 36th International Conference on
  Computer-Aided Design}, ICCAD ’17, page 131–138. IEEE Press, 2017.

\bibitem{Drone}
D.~Palossi.
\newblock Ultra low power deep-learning-powered autonomous nano drones.
\newblock {\em CoRR}, abs/1805.01831, 2018.

\bibitem{momentum}
I.~Sutskever~et al.
\newblock On the importance of initialization and momentum in deep learning.
\newblock In {\em ICML 2013}, ICML’13, page III–1139–III–1147.
  JMLR.org, 2013.

\bibitem{BB_target}
Y.~Liu et~al.
\newblock Delving into transferable adversarial examples and black-box attacks.
\newblock {\em CoRR}, abs/1611.02770, 2016.

\bibitem{advTraining}
I.~J.~Goodfellow et~al.
\newblock Explaining and harnessing adversarial examples, 2014.

\bibitem{FGSM}
Ian~J. Goodfellow, Jonathon Shlens, and Christian Szegedy.
\newblock Explaining and harnessing adversarial examples, 2014.

\bibitem{Madry}
A.~Madry et~al.
\newblock Towards deep learning models resistant to adversarial attacks, 2017.

\bibitem{adv_defense}
C.~Guo et~al.
\newblock Countering adversarial images using input transformations.
\newblock {\em CoRR}, abs/1711.00117, 2017.

\bibitem{TF}
Mart\'{\i}n~Abadi et~al.
\newblock {TensorFlow}: Large-scale machine learning on heterogeneous systems,
  2015.
\newblock Software available from tensorflow.org.

\bibitem{defenseGAN}
P.~Samangouei et~al.
\newblock Defense-gan: Protecting classifiers against adversarial attacks using
  generative models.
\newblock {\em CoRR}, abs/1805.06605, 2018.

\bibitem{var_autoenc}
U.~{Hwang}, J.~{Park}, H.~{Jang}, S.~{Yoon}, and N.~I. {Cho}.
\newblock Puvae: A variational autoencoder to purify adversarial examples.
\newblock {\em IEEE Access}, 7:126582--126593, 2019.

\bibitem{empir}
Sanchari Sen, Balaraman Ravindran, and Anand Raghunathan.
\newblock {EMPIR}: Ensembles of mixed precision deep networks for increased
  robustness against adversarial attacks.
\newblock In {\em International Conference on Learning Representations}, 2020.

\bibitem{ensem_fp}
Tianyu Pang, Kun Xu, Chao Du, Ning Chen, and Jun Zhu.
\newblock Improving adversarial robustness via promoting ensemble diversity.
\newblock {\em CoRR}, abs/1901.08846, 2019.

\bibitem{mnist}
Yann LeCun and Corinna Cortes.
\newblock {MNIST} handwritten digit database.
\newblock 2010.

\bibitem{imagenet}
J.~Deng, W.~Dong, R.~Socher, L.-J. Li, K.~Li, and L.~Fei-Fei.
\newblock {ImageNet: A Large-Scale Hierarchical Image Database}.
\newblock In {\em CVPR09}, 2009.

\bibitem{AllConvNet}
J.T.~Springenberg et~al.
\newblock Striving for simplicity: The all convolutional net.
\newblock In {\em ICLR (workshop track)}, 2015.

\end{thebibliography}


\end{document}